%% file: iclr2025_conference.tex
\documentclass{article} 
\usepackage{iclr2025_conference,times}

\input{math_commands.tex}

\usepackage{tikz}
\usepackage{pgfplots}
\usepackage{hyperref}
\usepackage{url}
\usepackage{booktabs}
\usepackage{graphicx}
\usepackage{subcaption}

\title{Karush-Kuhn-Tucker Condition-Trained Neural Networks (KKT Nets)\thanks{After independently conceiving the idea of finding the primal and dual optimal solutions by minimizing the KKT condition-related losses in a neural network that takes problem parameters as input, we were actively working on the results when we became aware of similar work \cite{femine2024kktinformedneuralnetwork}, which was released after we began working on the idea. 
The code associated with our work is available at: \url{https://github.com/Shreya-a-a/KKTNetworks}.}}

\author{Shreya Arvind,  Rishabh Pomaje \& Rajshekhar V. Bhat \thanks{All the authors contributed equally to the work.}  \\
Department of Electrical, Electronics and Communication Engineering\\
Indian Institute of Technology Dharwad\\
Karnataka, India\\
\texttt{\{210020046,210020036,rajshekhar.bhat\}@iitdh.ac.in} \\
}

\iclrfinalcopy 
\begin{document}

\maketitle

\begin{abstract}
This paper presents a novel approach to solving convex optimization problems by leveraging the fact that, under certain regularity conditions, any set of primal or dual variables satisfying the Karush-Kuhn-Tucker (KKT) conditions is necessary and sufficient for optimality. Similar to Theory-Trained Neural Networks (TTNNs), the parameters of the convex optimization problem are input to the neural network, and the expected outputs are the optimal primal and dual variables. A choice for the loss function in this case is a loss, which we refer to as the KKT Loss, that measures how well the network's outputs satisfy the KKT conditions. We demonstrate the effectiveness of this approach using a linear program as an example. For this problem, we observe that minimizing the KKT Loss alone outperforms training the network with a weighted sum of the KKT Loss and a Data Loss (the mean-squared error between the ground truth optimal solutions and the network’s output). Moreover, minimizing only the Data Loss yields inferior results compared to those obtained by minimizing the KKT Loss. While the approach is promising, the obtained primal and dual solutions are not sufficiently close to the ground truth optimal solutions. In the future, we aim to develop improved models to obtain solutions closer to the ground truth and extend the approach to other problem classes.
\end{abstract}

\section{Introduction}
Recently, there has been growing interest in utilizing the deep learning framework to solve optimization problems. In this work, we present a neural network-based approach that leverages the Karush-Kuhn-Tucker (KKT) conditions to approximately solve convex optimization problems.

The general form of a convex optimization problem, \cite{Boyd_Vandenberghe_2004}, is expressed as:
\begin{subequations}
\begin{align}
\min_{\vx \in \mathbb{R}^n} \quad & f_0(\vx), \\
\text{subject to} \quad & f_i(\vx)  \leq 0, \quad i = 1, \ldots, m, \\
& g_i(\vx) = 0, \quad i = 1, \ldots, p, 
\end{align}
\end{subequations}
where $\vx = [x_1, x_2, \ldots, x_n] \in \mathbb{R}^n$, $f_i: \mathbb{R}^n \rightarrow \mathbb{R}$ are convex functions and  $g_i: \mathbb{R}^n \rightarrow \mathbb{R}$ are affine. The domain of the above problem is defined as: $\mathcal{D} = \bigcap_{i=0}^{m} \text{\textbf{dom}} f_i\ \cap\ \bigcap_{i=1}^{p} \text{\textbf{dom}} g_i$. A dual formulation of the above optimization problem is derived using the Lagrangian, where the Lagrangian,  $\mathcal{L}: \mathbb{R}^n \times \mathbb{R}^m \times \mathbb{R}^p \rightarrow \mathbb{R}$ is defined as, 
\begin{align}
   \mathcal{L}(\vx, \boldsymbol{\lambda}, \boldsymbol{\nu}) = f_0(\vx) + \sum_{i=1}^{m} \lambda_i f_i(\vx) + \sum_{i=1}^{p} \nu_i g_i(\vx), 
\end{align} 
for  $\boldsymbol{\lambda} = [\lambda_1, \lambda_2, \ldots, \lambda_m] \in \mathbb{R}^m$ and $\boldsymbol{\nu} = [\nu_1, \nu_2, \ldots, \nu_p] \in \mathbb{R}^p$, 
where \( \lambda_i\geq 0 \) is the dual variable associated with the inequality constraint \( f_i(x) \) for \( i \in \{1,2,\ldots,m\} \), and \( \nu_i \) is the dual variable associated with the equality constraint \( g_i(x) \) for \( i \in \{1,2,\ldots,p\} \).

The  Lagrange dual function, denoted by $g(\boldsymbol{\lambda}, \boldsymbol{\nu})$, is given as,
\begin{align}
   g(\boldsymbol{\lambda}, \boldsymbol{\nu}) = \inf_{\vx \in \mathcal{D}} \mathcal{L}(\vx, \boldsymbol{\lambda}, \boldsymbol{\nu}).
\end{align}


In convex optimization problems, under regularity conditions such as Slater's condition for inequality constraints, the following KKT conditions provide necessary and sufficient conditions for optimality:
\begin{subequations}\label{eq:kkt_conditions}
    \begin{align}
        &\text{Primal feasibility} &&: f_i(\vx^*) \leq 0, \quad i = 1, \ldots, m, \\&{} &&: g_i(\vx^*) = 0, \quad i = 1, \ldots, p, \label{eq:primal_feasibility}\\
        &\text{Dual feasibility} &&: \lambda_i^* \geq 0, \quad i = 1, \ldots, m, \label{eq:dual_feasibility}\\
        &\text{Complementary slackness} &&: \lambda_i^* f_i(\vx^*) = 0, \quad \quad i = 1, \ldots, m, \quad \text{and, }\label{eq:complementary_slackness}\\
        &\text{Stationarity} &&: \nabla f_0(\vx^*) + \sum_{i=1}^{m} \lambda_i^* \nabla f_i(\vx^*) + \sum_{i=1}^{p} \nu_i^* \nabla g_i(\vx^*) = 0, \label{eq:stationarity}
    \end{align}
\end{subequations}
where \( \vx^* \) is the optimal primal variable and \( \{ \lambda_i^* \}_{i=1}^m \) and \( \{ \nu_i^* \}_{i=1}^p \) are the optimal dual variables. Here, \( \nabla f_i(\vx^*) \) represents the gradient of \( f_i(\vx) \) evaluated at \( \vx = \vx^* \).
 
In this article, we focus on training a deep learning model to take problem parameters for specific subclasses of convex optimization problems, such as linear programs, as inputs and output the optimal primal variable \( \vx^* \) and dual variables $\boldsymbol{\lambda}^*$ and $\boldsymbol{\nu}^* $.  We present an architecture that embeds the KKT conditions for optimality into the neural network and defines loss functions accordingly. We refer to these networks as Karush-Kuhn-Tucker Condition-Trained Neural Networks, abbreviated as KKT Nets.

\section{Neural Network Approach to Solving Convex Optimization Problems} 
\label{sec:section_2}
As mentioned, for a convex optimization problem, if we find $\vx^*$, $\boldsymbol{\lambda^*}$ and $\boldsymbol{\nu^*}$ that satisfy \eqref{eq:primal_feasibility} - \eqref{eq:stationarity}, they must be the optimal primal and dual solutions. In our approach, we take the parameters of a problem as input, and the expected output is $\vx^*$, $\boldsymbol{\lambda^*}$ and $\boldsymbol{\nu^*}$.  Our loss function includes what we refer to as KKT Loss \( (L_{\rm KKT}) \), which is a weighted sum of the primal feasibility loss, dual feasibility loss, complementary slackness loss, and stationarity loss, where
\begin{align}
\text{Primal Feasibility Loss}, &\quad L_{\rm PF} = \frac{1}{m} \sum_{i=1}^{m} \max(0, f_i(\hat{\vx}))^2, \\
\text{Dual Feasibility Loss}, &\quad L_{\rm DF} = \frac{1}{m} \sum_{i=1}^{m} \max(0, -\hat{\lambda}_i)^2, \\
\text{Complementary Slackness Loss}, &\quad L_{\rm CS} = \frac{1}{m} \sum_{i=1}^{m} (\hat{\lambda}_i \cdot f_i(\hat{\vx}))^2, \\
\text{Stationarity Loss}, &\quad L_{\rm S} = \frac{1}{n}\left\|\nabla f_0(\hat{\vx}) + \sum_{i=1}^{m} \hat{\lambda}_i \nabla f_i(\hat{\vx}) + \sum_{j=1}^{p} \hat{\nu}_j \nabla g_j(\hat{\vx})\right\|_2^2,
\end{align}
where $\hat{\vx}$, $\hat{\boldsymbol{\lambda}}$, and $\hat{\boldsymbol{\nu}}$ represent the neural network outputs corresponding to $\vx^*$, $\boldsymbol{\lambda^*}$ and $\boldsymbol{\nu^*}$, respectively, and $\left\|.\right\|_2$ represents the 2-norm of a vector. Concretely,  the KKT Loss is defined as follows:
\begin{align}
    L_{\rm KKT}(\alpha_1, \alpha_2, \alpha_3, \alpha_4) = \alpha_1 L_{\rm PF} + \alpha_2 L_{\rm DF} + \alpha_3 L_{\rm CS} + \alpha_4 L_{\rm S},
\end{align}
where \(\alpha_1, \ldots, \alpha_4\) are non-negative weights that may be treated as hyperparameters.

In addition, we can achieve optimal results for convex optimization problems using frameworks such as CVX. However, our goal is to enable neural networks to learn to output the optimal solution. To this end, we incorporate both the KKT Loss and the mean-squared error between the optimal solution and the solution output by the neural network, referred to as Data Loss, as additional loss functions during the training phase. Concretely, 
\begin{equation}
\text{Data Loss}, L_{\rm Data} = \frac{1}{K} \sum_{k=1}^{K} \left\|\vy^*_{k} - \hat{\vy}_k\right\|^2,
\end{equation}
where $\vy^*_{k}$ and $\hat{\vy_k}$ represent the ground truth and the solution output by the neural network, i.e., $ \vy^*_k =  [\vx^*, \boldsymbol{\lambda}^*, \boldsymbol{\nu}^*] $ and $ \hat{\vy}_k = [\hat{\vx}, \hat{\boldsymbol{\lambda}}, \hat{\boldsymbol{\nu}}]$ for the $k^{\rm th}$ example, where $k \in \{1, 2, \ldots, K\}$ and $K$ is the total number of examples.
The combined loss function is a weighted sum of the KKT Loss and the Data Loss:
\begin{align}\label{eq-closs}
\text{Combined Loss}, \quad L(\alpha_1, \alpha_2, \alpha_3, \alpha_4, \beta)  = L_{\rm KKT}(\alpha_1, \alpha_2, \alpha_3, \alpha_4)  + \beta L_{\rm Data}, 
\end{align}
where $\alpha_1, \ldots, \alpha_4$ and $\beta$ are hyperparameters that determine the weighting of different losses. We explore how the model behaves under various values of these hyperparameters. When $\beta = 0$, only the KKT Loss is considered; when $\alpha_1, \ldots, \alpha_4 = 0$, only the Data Loss is used.

\section{Dataset Generation, Model Training, and Results}
In this section, we present the dataset generation and preparation, along with the training of a neural network model and the results obtained. Note that to minimize only the KKT Loss, we need only the model parameters and do not require the ground truth optimal solutions. However, to minimize the Data Loss and evaluate performance, we need the ground truth optimal solution for the given model parameters.

\subsection{Data Set Generation}
In this work, we consider a class of optimization problems that can be expressed with explicit, closed-form expressions, referred to as \emph{parameterized problems}.

For training a neural network (in our case, to minimize the Data Loss), we require labeled data, specifically the problem parameters and the corresponding optimal primal and dual solutions.  We can artificially generate this data, which consists of the parameters of a problem instance and the corresponding solutions. To achieve this, we can use random number generators to populate the parameters of optimization problems. Furthermore, to find the optimal primal and dual solutions, one can use any of the numerous available solvers. Specifically, for our data, we used the CVXPY, \cite{agrawal2018rewriting, diamond2016cvxpy}.
Below, we explain how the data can be generated using the standard form of a quadratic programming (QP) problem, which is as follows:
\begin{subequations}
   \begin{align}
      \min_{\vx} \quad &\frac{1}{2}\vx^T \mP \vx + \vq^T \vx + r, \\
      \text{subject to} \quad &\mG \vx \preceq \vh, \\
      \quad &\mA \vx = \vb,  
   \end{align}
\end{subequations}
where $\mP \in \mathbb{R}^{n \times n}$, $\vq \in \mathbb{R}^{n}$, $r \in \mathbb{R}$, $\mG \in \mathbb{R}^{m \times n}$, $\vh \in \mathbb{R}^{m}$, $\mA \in \mathbb{R}^{p \times n}$, $\vb \in \mathbb{R}^{p}$ are the parameters of the problem that we take as inputs to the neural network.


We use a random number generator to populate the entries of the matrices and vectors (of parameters) in the above expression. 
We then normalize the entries in the matrices and vectors to the interval $[-1, 1]$, as described below. Let  $\Theta = \max\{P_{\max}, q_{\max}, r, G_{\max}, h_{\max}, A_{\max}, b_{\max}\}$,
where each element of the set above is the maximum absolute value of the entries of the corresponding matrix or vector. For example,
$G_{\max} \overset{\Delta}{=} \max\{|g_{ij}|\}$, where $g_{ij}$ is the element in the $i^{\text{th}}$ row and $j^{\text{th}}$ column. We then perform the complete normalization as follows: 
$\tilde{\mP} = \mP / \Theta$, $\tilde{\vq} = \vq / \Theta$, $\tilde{r} = r / \Theta$, $\tilde{\mG} = \mG / \Theta$, $\tilde{\vh} = \vh / \Theta$, $\tilde{\mA} = \mA / \Theta$, and $\tilde{\vb} = \vb / \Theta$.
The normalized problem will be of the form: 
\begin{subequations}
   \begin{align}
      \min_{\vx} \quad &\frac{1}{2}\vx^T \tilde{\mP} \vx + \tilde{\vq}^T \vx + \tilde{r}, \\
      \text{subject to} \quad &\tilde{\mG} \vx \preceq \tilde{\vh}, \\
      \quad &\tilde{\mA} \vx = \tilde{\vb}.   
   \end{align}
\end{subequations}
If the solution to the normalized problem is $\tilde{\vx}^*$, then due to the non-negative uniform scaling, we get the solution to the original problem as, 
${\vx}^* = \tilde{\vx}^*$.

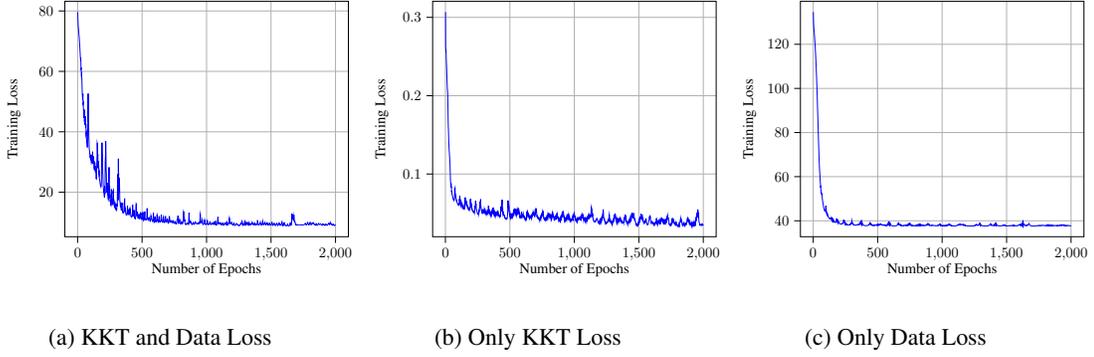
\begin{figure}[t]
    \centering
    \begin{subfigure}{0.3\textwidth}
        \centering
        \input{kktandmseplot.tex} 
        \caption{KKT and Data Loss}
        \label{fig:kktandmse}
    \end{subfigure}
    \hfill
    \begin{subfigure}{0.3\textwidth}
        \centering
        \input{nomseplot.tex} 
        \caption{Only KKT Loss}
        \label{fig:nomse}
    \end{subfigure}
    \hfill
    \begin{subfigure}{0.3\textwidth}
        \centering
        \input{onlymseplot.tex} 
        \caption{Only Data Loss}
        \label{fig:onlymse}
    \end{subfigure}
    \caption{Evolution of the training losses over the number of epochs when the network is trained to minimize different combinations of KKT and Data Losses.}
    \label{fig:trainingloss}
\end{figure}

\subsection{Training the Network}
While the approach has the potential to work for any parameterized convex optimization problem, we consider the following simple linear optimization problem and generate the dataset:
\begin{subequations}
\begin{align}
   \min_{\vx \in \mathbb{R}^2} \quad & \vc^T \vx, \\
   \text{subject to} \quad & \mA \vx \leq \vb, 
\end{align}
\end{subequations}
where $\mA \in \mathbb{R}^{2\times 2}$, $\vc \in \mathbb{R}^2$, and $\vb \in \mathbb{R}^2$ are the parameters of the problem, the flattened versions of which are taken as inputs to the neural network.  
As mentioned, we used CVXPY to generate instances of these problems, where the elements in $\mA$, $\vc$, and $\vb$ were randomly generated. Each problem was normalized and solved to obtain the primal and dual solutions. The resulting dataset, consisting of the coefficient matrices along with the primal and dual solutions, was used to train the network.   Only those problems that were feasible and resulted in optimal, accurate solutions were considered for training purposes.

The network was trained using three different loss configurations: only Data Loss, with $\alpha_1 = \alpha_2 = \alpha_3 = \alpha_4 = 0$ and $\beta = 1$; only KKT Loss with $\alpha_1 = 0.1$, $\alpha_2 = 0.1$, $\alpha_3 = 0.2$, $\alpha_4 = 0.6$, and $\beta = 0$; and a combination of KKT and Data Loss, with $\alpha_1 = 0.1$, $\alpha_2 = 0.1$, $\alpha_3 = 0.2$, $\alpha_4 = 0.6$, and $\beta = 1$. 

\subsection{Results}
Fig.~\ref{fig:trainingloss} shows the decrease in training loss when different loss functions are considered for the KKT Net. We observe that, regardless of which loss is used, the model demonstrates its ability to learn, as indicated by the reduction in losses throughout the training process.

We next present results for the inference done on the trained models on an independently generated dataset, which is normalized before using it for inference similar to what is done during training. 

The root-mean-square error (RMSE) between the primal and dual solutions output by the trained networks—each trained to minimize different combinations of KKT and Data Loss—and the ground truth optimal solutions obtained using CVXPY is presented in Table~\ref{tab:rmse_table}. We observe that the performance of the KKT Net, when trained to minimize different combinations of KKT and Data Loss functions, is nearly identical, with the case of minimizing only the KKT loss performing slightly better than the others. 

\begin{figure}[t]
    \centering
    \begin{subfigure}{0.35\textwidth}
        \centering
        \includegraphics[width=\textwidth]{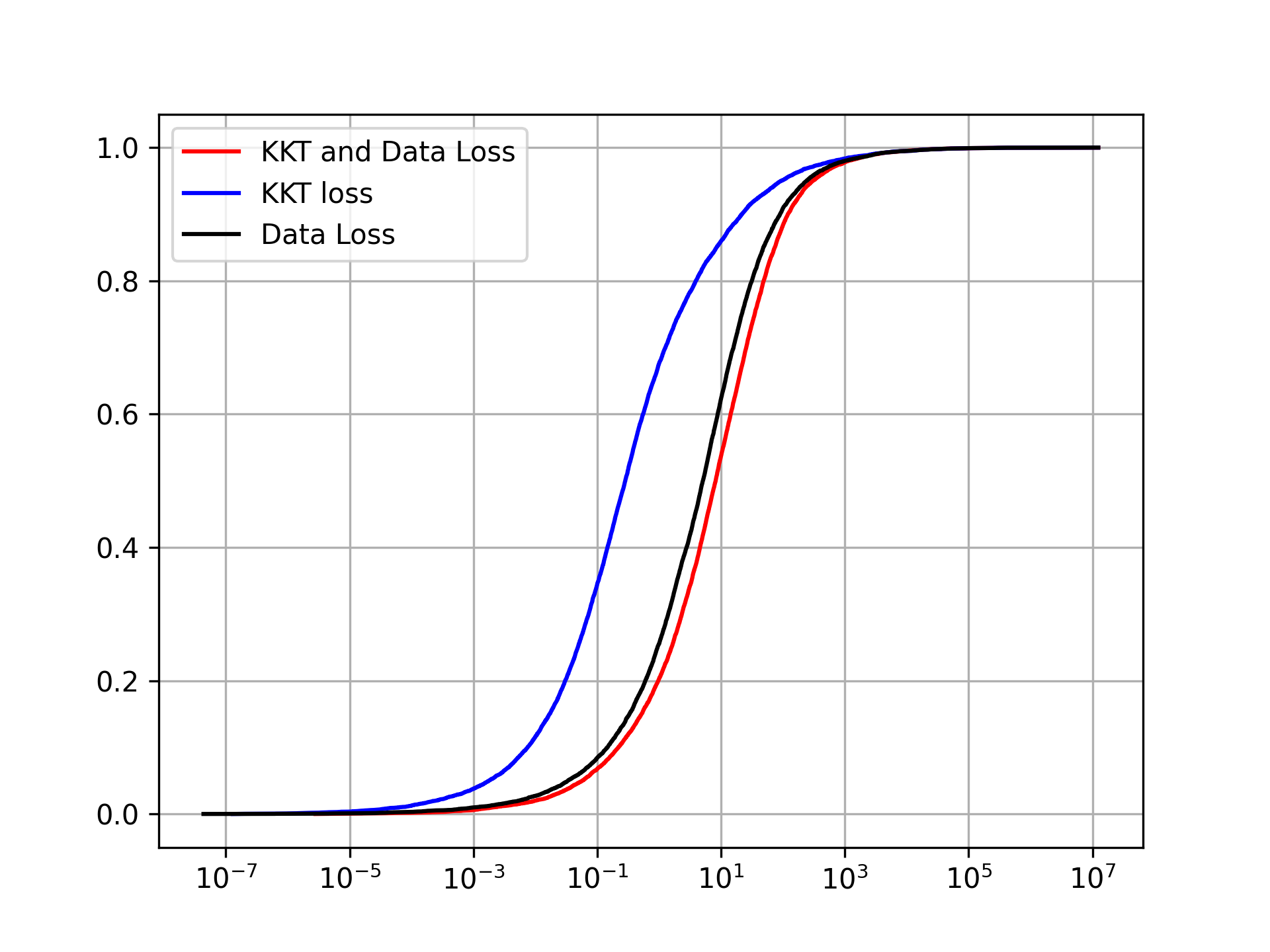}  
        \caption{CDF of $(\hat{x_1} - x_1^*)^2$}
        \label{fig:cdf_x1}
    \end{subfigure}
    \hfill
    \begin{subfigure}{0.35\textwidth}
        \centering
        \includegraphics[width=\textwidth]{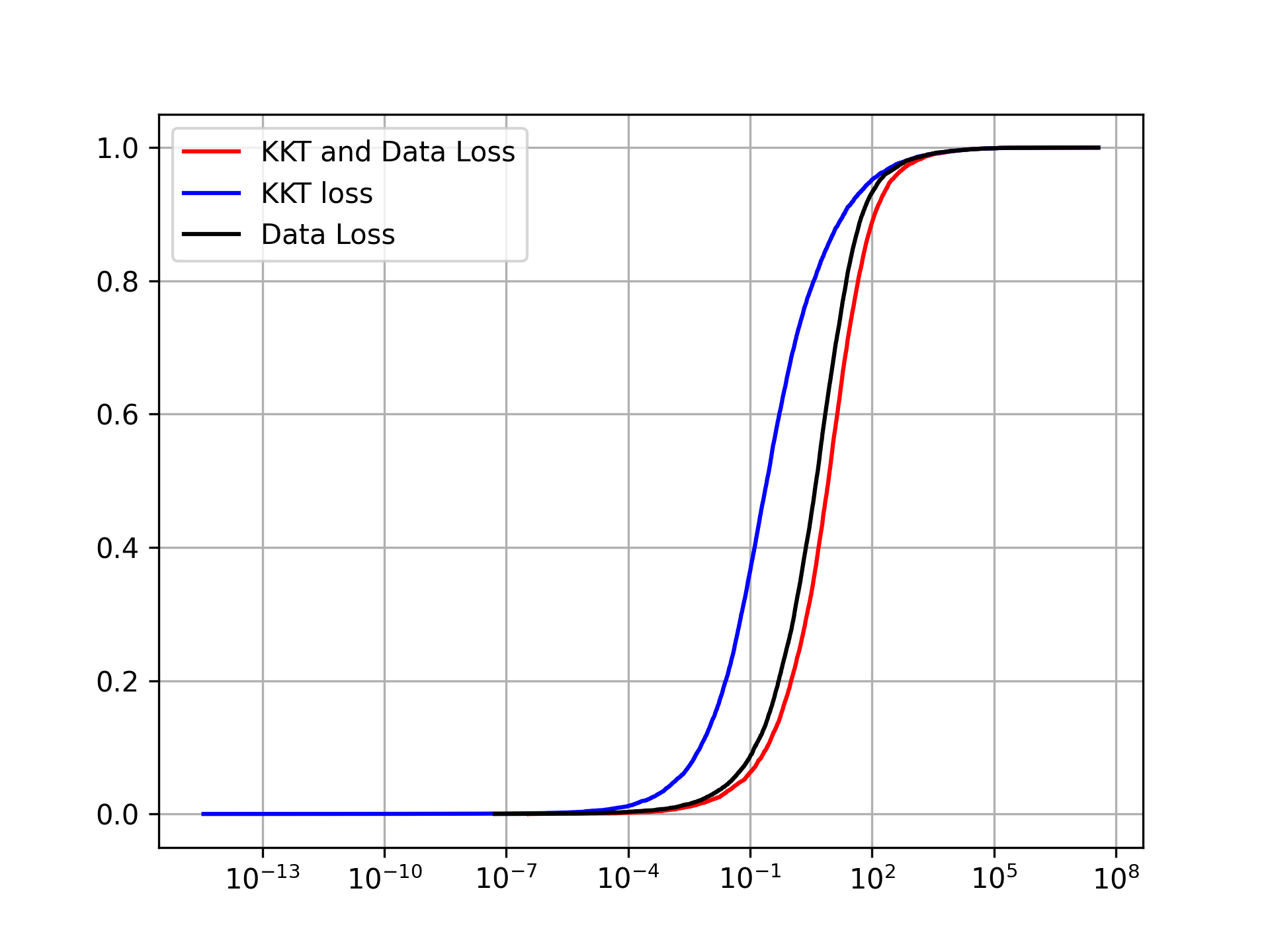}  
        \caption{CDF of $(\hat{x_2} - x_2^*)^2$}
        \label{fig:cdf_x2}
    \end{subfigure}
    
    \vspace{0.5cm}  
    
    \begin{subfigure}{0.35\textwidth}
        \centering
        \includegraphics[width=\textwidth]{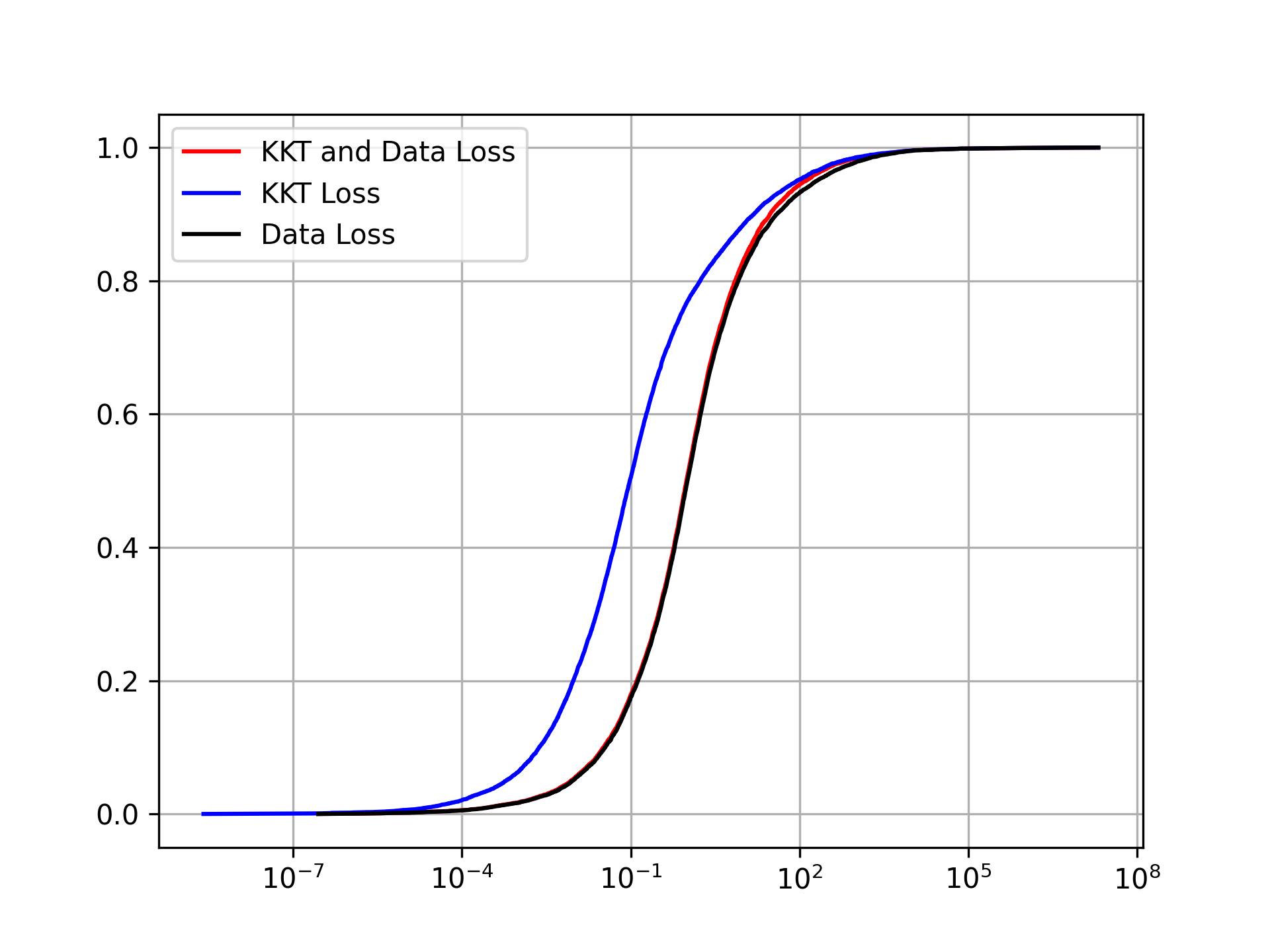}  
        \caption{CDF of $(\hat{\lambda_1} - \lambda_1^*)^2$}
        \label{fig:cdf_lambda1}
    \end{subfigure}
    \hfill
    \begin{subfigure}{0.35\textwidth}
        \centering
        \includegraphics[width=\textwidth]{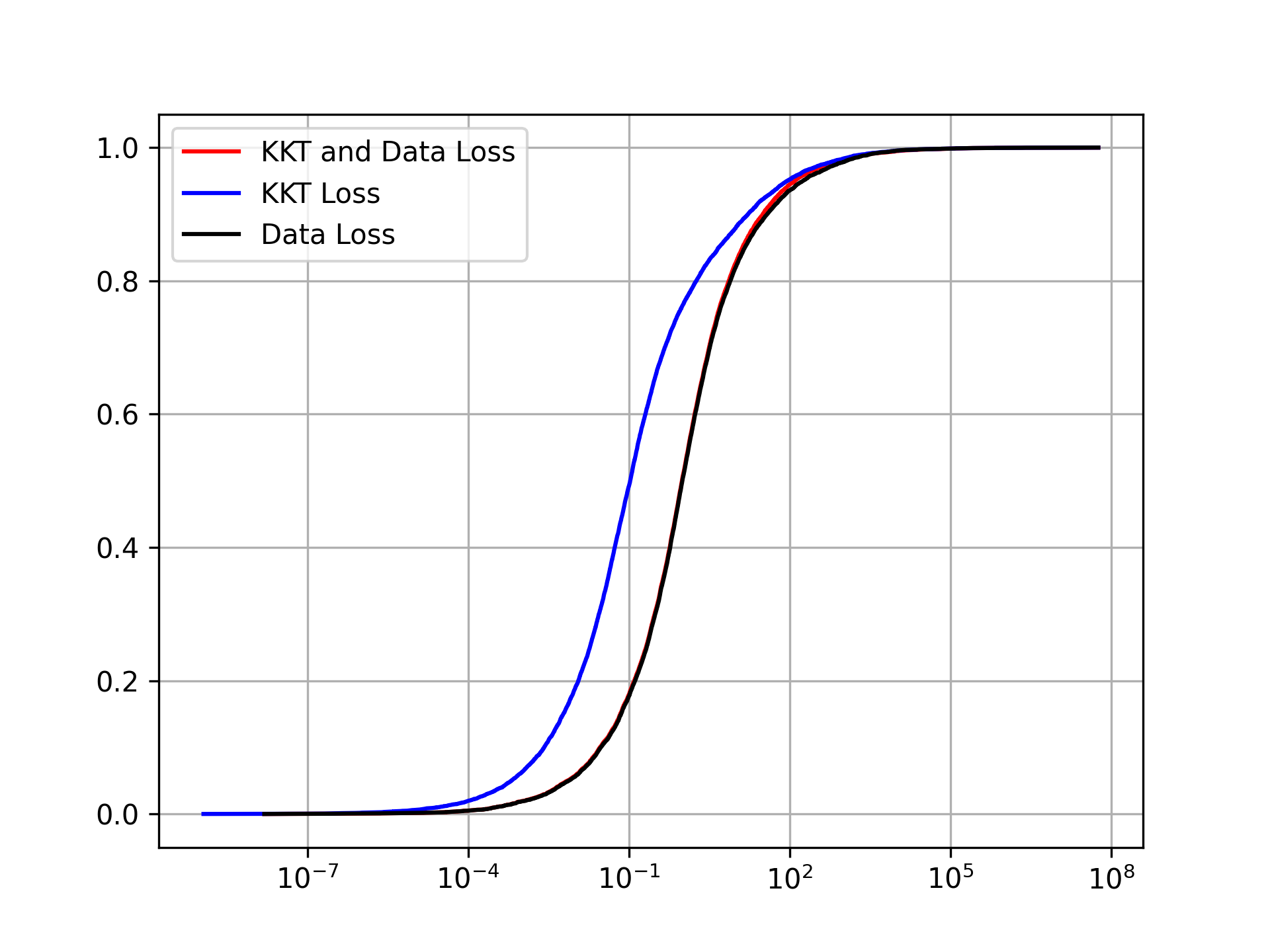}  
        \caption{CDF of $(\hat{\lambda_2} - \lambda_2^*)^2$}
        \label{fig:cdf_lambda2}
    \end{subfigure}
    
    \caption{CDFs of squared errors between the primal and dual solutions output by the KKT Net and the ground truth optimal solutions obtained using CVXPY.
    }
    \label{fig:CDF}
\end{figure}

Given that the performance differences are not significant, it may be instructive to examine the cumulative distribution functions (CDFs) of the squared errors between the optimal primal and dual solutions output by the KKT Net and the ground truth solutions obtained using CVXPY, as shown in Fig.~\ref{fig:CDF}. From the figure, we note that the network performs best when only \( L_{\rm KKT} \) is minimized, exhibiting a higher proportion of smaller errors compared to when \( L_{\rm Data} \) and \( L_{\rm KKT} + L_{\rm Data} \) are minimized.

\begin{table}[t]
    \centering
    \begin{tabular}{lccc}
        \toprule
        & Only Data Loss & Only KKT Loss & KKT and Data Loss \\
        \midrule
        $\hat{x_1}$       & 47.946          & 47.672         & 48.050            \\
        $\hat{x_2}$       & 65.683          & 65.692         & 65.883            \\
        $\hat{\lambda_1}$ & 66.231          & 66.078         & 66.199             \\
        $\hat{\lambda_2}$ & 98.845          & 98.904         & 99.195              \\
        \bottomrule
    \end{tabular}
    \caption{The  RMSE between the primal and dual solutions output by the trained networks—each trained to minimize different combinations of KKT and Data Loss—and the ground truth optimal solutions obtained using CVXPY.}
    \label{tab:rmse_table}
\end{table}

\section{Conclusion}
In this paper, we presented an approach to solving convex optimization problems using a neural network, where the input consisted of the problem parameters and the expected output was the optimal primal and dual variables. We formulated the problem to minimize the KKT Loss, which measures how closely the solution output by the neural network satisfies the KKT conditions. Additionally, we considered a combined loss, defined as a weighted sum of the KKT Loss and what we referred to as Data Loss, which is the MSE between the ground truth optimal primal and dual variables and those predicted by the neural network. We used a simple linear program to evaluate the performance of this approach and found that the neural network was able to learn to output the optimal primal and dual solutions.  Training the network to minimize only the KKT Loss resulted in better performance, with a larger fraction of smaller errors compared to training it to minimize the combined loss or only the Data Loss. While the approach is promising, the obtained primal and dual solutions are not sufficiently close to the ground truth optimal solutions. In future work, we aim to develop improved models to obtain solutions that are closer to the ground truth and to extend the approach to other classes of convex and non-convex optimization problems.

\bibliography{iclr2025_conference}
\bibliographystyle{iclr2025_conference}


\end{document}

%% file: math_commands.tex
\usepackage{amsmath,amsfonts,bm}

\def\eqref#1{equation~\ref{#1}}

\def\1{\bm{1}}

\def\vb{{\bm{b}}}
\def\vc{{\bm{c}}}

\def\vh{{\bm{h}}}

\def\vq{{\bm{q}}}

\def\vx{{\bm{x}}}
\def\vy{{\bm{y}}}

\def\mA{{\bm{A}}}

\def\mG{{\bm{G}}}

\def\mP{{\bm{P}}}

\DeclareMathAlphabet{\mathsfit}{\encodingdefault}{\sfdefault}{m}{sl}
\SetMathAlphabet{\mathsfit}{bold}{\encodingdefault}{\sfdefault}{bx}{n}



%% file: kktandmseplot.tex
\begin{tikzpicture}[scale = 0.55]

\definecolor{darkgray176}{RGB}{176,176,176}

\begin{axis}[
tick align=outside,
tick pos=left,
x grid style={darkgray176},
xlabel={Number of Epochs},
xmajorgrids,
xmin=-98.95, xmax=2099.95,
xtick style={color=black},
y grid style={darkgray176},
ylabel={Training Loss},
ymajorgrids,
ymin=5.28232205708822, ymax=83.1739208221435,
ytick style={color=black}
]
\addplot [semithick, blue]
table {%
1 79.6333936055501
2 77.9018700917562
3 76.8113562266032
4 75.6462287902832
5 74.8642222086589
6 74.3919359842936
7 73.914021174113
8 73.3940302530925
9 72.9162089029948
10 72.4224414825439
11 71.9245389302572
12 71.3929926554362
13 70.7915884653727
14 70.159454981486
15 69.4866053263346
16 68.7733300526937
17 68.0703035990397
18 67.24578221639
19 66.4249871571859
20 65.6034984588623
21 64.8035456339518
22 64.2319717407227
23 64.0335070292155
24 64.2250734965007
25 64.12464650472
26 61.4016679128011
27 60.8186480204264
28 60.897611618042
29 58.7204742431641
30 58.4690500895182
31 58.7007891337077
32 57.7889919281006
33 57.4437700907389
34 55.6423778533936
35 54.0831057230631
36 52.9110921223958
37 52.1230634053548
38 51.3364105224609
39 51.1799996693929
40 51.650074005127
41 52.5303287506104
42 49.9869263966878
43 49.3447278340658
44 49.4906355539958
45 47.6803385416667
46 47.0457903544108
47 46.0987822214762
48 45.3321660359701
49 45.1883176167806
50 44.5440966288249
51 47.3476448059082
52 46.9403076171875
53 43.3545392354329
54 44.4234097798665
55 44.0892353057861
56 43.1772486368815
57 44.6414279937744
58 42.820810953776
59 42.1925913492839
60 44.5942783355713
61 45.050864537557
62 42.3673775990804
63 41.4147926966349
64 42.6662368774414
65 41.101634979248
66 39.2840741475423
67 39.5134983062744
68 38.7900387446086
69 37.6069920857747
70 37.142983118693
71 36.581782023112
72 36.2637278238932
73 35.6712538401286
74 35.4152062733968
75 34.8982963562012
76 34.9274816513062
77 35.6002330780029
78 40.8244867324829
79 42.2077643076579
80 44.3550662994385
81 46.4537925720215
82 52.6502564748128
83 43.3554242451986
84 43.8027801513672
85 43.29754002889
86 40.7641264597575
87 38.2515977223714
88 37.2071266174316
89 36.318603515625
90 34.3156112035116
91 34.1582562128703
92 33.8518250783285
93 33.2355209986369
94 32.8228486378988
95 32.2590796152751
96 31.9501927693685
97 31.6802504857381
98 31.7407398223877
99 31.377649307251
100 31.1896003087362
101 31.1645940144857
102 31.3995246887207
103 30.967493057251
104 30.5900932947795
105 31.0437920888265
106 32.1520150502523
107 32.0653120676676
108 30.9274749755859
109 30.9366505940755
110 30.3322105407715
111 30.0930910110474
112 30.5217138926188
113 31.6281318664551
114 29.7106965382894
115 31.7936579386393
116 33.4291763305664
117 32.0271650950114
118 29.3700294494629
119 30.6738758087158
120 31.9737841288249
121 30.7393658955892
122 30.533442179362
123 30.6954917907715
124 31.6168257395426
125 30.5070088704427
126 28.8714679082235
127 28.2818307876587
128 28.2624438603719
129 27.5345726013184
130 27.4654280344645
131 28.2672268549601
132 28.8011178970337
133 29.7900104522705
134 29.6817557017008
135 28.0966739654541
136 27.9256916046143
137 28.5680777231852
138 27.4630470275879
139 26.462366104126
140 26.8407417933146
141 26.6391331354777
142 26.4809703826904
143 25.2777862548828
144 24.827169418335
145 24.8571440378825
146 24.3024218877157
147 25.298090616862
148 28.3871825536092
149 29.6834875742594
150 30.9271055857341
151 32.4398632049561
152 36.2702407836914
153 36.0670789082845
154 34.6857007344564
155 32.5545069376628
156 32.0790278116862
157 32.5274378458659
158 28.559627532959
159 27.9357268015544
160 30.3345959981283
161 29.9085369110107
162 29.2101732889811
163 28.2178185780843
164 28.2806116739909
165 26.8561655680339
166 26.9758758544922
167 25.8616402943929
168 26.0283063252767
169 24.5677916208903
170 24.3119179407756
171 24.0093650817871
172 23.9308859507243
173 23.5372988382975
174 23.0230757395426
175 22.8375568389893
176 22.5502065022786
177 22.3219261169434
178 22.2577047348022
179 22.0681177775065
180 22.06383228302
181 21.8697770436605
182 22.0932366053263
183 22.3158384958903
184 22.7344175974528
185 24.5035966237386
186 30.520383199056
187 28.8902683258057
188 33.6281350453695
189 36.3995602925619
190 34.0451637903849
191 32.1638024648031
192 32.2018375396729
193 28.2618567148844
194 28.1931355794271
195 26.1372674306234
196 25.9005959828695
197 24.0361830393473
198 22.9013115564982
199 21.8377749125163
200 21.1344216664632
201 20.8092791239421
202 20.5905224482218
203 20.9218018849691
204 20.0879344940186
205 19.7287712097168
206 19.5226669311523
207 18.839516321818
208 18.3459040323893
209 20.574028968811
210 21.9664128621419
211 21.1818968454997
212 19.4277181625366
213 22.2970422108968
214 29.9709440867106
215 31.087210337321
216 32.1089089711507
217 34.2875423431396
218 36.8894894917806
219 35.0048147837321
220 32.1886285146077
221 28.4440364837646
222 26.4259853363037
223 26.1499989827474
224 23.0371602376302
225 21.4817571640015
226 20.0065962473551
227 21.1159474054972
228 23.1444164911906
229 22.2457799911499
230 21.8037668863932
231 20.4011869430542
232 19.8080981572469
233 19.1414213180542
234 18.8656352361043
235 18.5953667958577
236 18.2262001037598
237 17.9322363535563
238 17.5887149175008
239 17.7642084757487
240 18.6884377797445
241 22.9229526519775
242 24.1333287556966
243 27.4740384419759
244 28.2682317097982
245 22.751942952474
246 22.4776293436686
247 21.201410929362
248 20.5579493840535
249 20.7628644307454
250 18.6244230270386
251 18.7597411473592
252 18.2683067321777
253 16.9330348968506
254 17.366317431132
255 16.6044171651204
256 16.4352858861287
257 15.8092947006226
258 15.8188912073771
259 15.6940539677938
260 16.0920426050822
261 17.715944925944
262 20.8809611002604
263 20.8984800974528
264 19.350502649943
265 18.4109306335449
266 17.8051748275757
267 16.8005278905233
268 16.6906604766846
269 16.3779341379801
270 16.1032625834147
271 15.914587020874
272 15.9267196655273
273 16.2283935546875
274 17.0649766921997
275 19.8377672831217
276 19.7360407511393
277 20.7212715148926
278 20.8006207148234
279 17.6257298787435
280 17.0511798858643
281 16.9297739664714
282 16.1043532689412
283 15.6958424250285
284 15.5864302317301
285 15.3270346323649
286 15.3068981170654
287 15.1470387776693
288 15.4454793930054
289 15.5775731404622
290 16.3599551518758
291 16.0849825541178
292 15.8973093032837
293 15.2872231801351
294 14.8546956380208
295 14.5917380650838
296 14.656751314799
297 14.3888400395711
298 14.2804781595866
299 14.0967102050781
300 14.0082286198934
301 13.930100440979
302 13.9800942738851
303 14.2074076334635
304 14.7515068054199
305 16.0802421569824
306 15.801659266154
307 15.4089428583781
308 17.2019869486491
309 21.6531162261963
310 23.8197717666626
311 25.904198328654
312 23.0816446940104
313 21.1819744110107
314 17.3998244603475
315 18.5087067286174
316 31.1106204986572
317 25.1651865641276
318 26.3141562143962
319 23.3376280466716
320 21.2885144551595
321 19.8480040232341
322 23.6041199366252
323 24.8027315139771
324 22.3778527577718
325 22.9178956349691
326 20.2345237731934
327 18.0036595662435
328 16.8257373174032
329 16.7233750025431
330 16.5571104685466
331 15.8167521158854
332 15.9953648249308
333 15.6868292490641
334 15.5253245035807
335 15.3401807149251
336 14.9040257136027
337 14.4728851318359
338 14.2684342066447
339 14.0404192606608
340 14.05198097229
341 14.1499446233114
342 14.7893985112508
343 15.4005047480265
344 16.5198396046956
345 14.7459195454915
346 14.4458939234416
347 14.2798515955607
348 13.7161823908488
349 13.5661958058675
350 13.671212832133
351 13.4330342610677
352 13.2388928731283
353 13.2624848683675
354 13.2042242685954
355 13.1800956726074
356 13.2215356826782
357 13.1685962677002
358 13.2421960830688
359 13.658434232076
360 15.1274598439535
361 15.9252109527588
362 16.3748594919841
363 15.9791405995687
364 17.7162771224976
365 17.7713813781738
366 15.5301678975423
367 14.3169342676798
368 14.4593944549561
369 13.7240238189697
370 13.6792805989583
371 13.7812865575155
372 13.5921904246012
373 13.3191134134928
374 13.0074055989583
375 12.7688814798991
376 12.5190658569336
377 12.5202821095785
378 12.6082105636597
379 12.5391124089559
380 12.5558563868205
381 12.6318171819051
382 12.7004585266113
383 12.9144503275553
384 13.2545719146729
385 14.0194918314616
386 14.7028363545736
387 15.5930147171021
388 14.1141490936279
389 13.9273662567139
390 14.016019821167
391 13.3242435455322
392 12.6747900644938
393 12.7534236907959
394 12.9766826629639
395 13.0921141306559
396 13.5350300470988
397 13.6635278065999
398 14.0800174077352
399 13.5362215042114
400 13.0403563181559
401 12.7897513707479
402 12.9164463678996
403 13.2043183644613
404 14.1082286834717
405 14.5513792037964
406 14.7222299575806
407 14.3623142242432
408 14.1746635437012
409 13.2400515874227
410 13.0900481541952
411 12.8856678009033
412 12.4648532867432
413 12.1088609695435
414 12.1216859817505
415 12.016547203064
416 11.8739312489827
417 11.8409903844198
418 11.7262239456177
419 11.7338205973307
420 11.7877178192139
421 11.8902708689372
422 12.0109764734904
423 12.2977492014567
424 12.7594645818075
425 13.7816410064697
426 14.9664154052734
427 15.8569533030192
428 14.6025142669678
429 13.0626490910848
430 12.5217615763346
431 12.6187324523926
432 12.5454371770223
433 12.2373797098796
434 12.1817607879639
435 12.7282342910767
436 13.3524312973022
437 14.1020669937134
438 13.5659863154093
439 13.0830958684285
440 12.7001126607259
441 12.0112380981445
442 11.5361792246501
443 11.5521815617879
444 11.4145962397257
445 11.3032954533895
446 11.1866156260173
447 11.1553738911947
448 11.2473704020182
449 11.5206232070923
450 12.0474227269491
451 13.405016263326
452 14.3751827875773
453 15.470588684082
454 16.0778754552205
455 15.9424187342326
456 13.5687859853109
457 12.9476280212402
458 13.3615897496541
459 12.7593110402425
460 12.1689872741699
461 12.2031151453654
462 12.7341445287069
463 13.5956214269002
464 12.7432905832926
465 11.9621464411418
466 11.7618662516276
467 11.5964539845785
468 11.3693669637044
469 11.3464714686076
470 11.3364191055298
471 11.4773817062378
472 11.5942443211873
473 11.8226973215739
474 12.1589247385661
475 12.7674967447917
476 12.7460826237996
477 12.5916938781738
478 12.0626691182454
479 11.743122736613
480 11.5163822174072
481 11.3063255945841
482 11.212984085083
483 11.3046712875366
484 11.3530950546265
485 11.5020780563354
486 11.6693836847941
487 12.0947831471761
488 12.6119677225749
489 13.3778553009033
490 12.8285697301229
491 12.7189776102702
492 12.7556082407633
493 12.0224129358927
494 11.3385712305705
495 11.3271907170614
496 11.2893489201864
497 11.2064453760783
498 11.0499741236369
499 10.9179134368896
500 10.9556436538696
501 11.127080599467
502 11.4088487625122
503 12.0019639333089
504 12.5121297836304
505 13.2791013717651
506 13.3541876475016
507 13.1742738087972
508 12.4595158894857
509 11.4275884628296
510 11.0548276901245
511 11.2756547927856
512 11.2849086125692
513 10.959210395813
514 10.7655407587687
515 10.8039782842
516 11.0976702372233
517 11.7029358545939
518 12.2143354415894
519 12.7826712926229
520 12.910808245341
521 13.6237875620524
522 12.728297551473
523 11.6748844782511
524 11.2487684885661
525 11.061816851298
526 10.9123477935791
527 10.9132258097331
528 10.8862454096476
529 10.9236895243327
530 10.9691591262817
531 10.9439004262288
532 10.8550504048665
533 10.8558928171794
534 10.9362834294637
535 11.0544341405233
536 11.3020849227905
537 12.0675020217896
538 13.2747551600138
539 14.6853275299072
540 14.0779174168905
541 13.1158806482951
542 11.7561054229736
543 11.3716548283895
544 11.2925497690837
545 11.3018999099731
546 11.0137090682983
547 10.7793413798014
548 10.8422854741414
549 11.1000506083171
550 11.237561861674
551 11.4818073908488
552 11.4681485493978
553 11.3334083557129
554 11.286262512207
555 11.5181891123454
556 11.3921070098877
557 11.2999458312988
558 10.8741537729899
559 10.7350231806437
560 10.659810702006
561 10.549420038859
562 10.4544808069865
563 10.6261259714762
564 10.8326718012492
565 10.997763633728
566 11.0152117411296
567 11.1491050720215
568 11.5401461919149
569 12.2453937530518
570 12.0589853922526
571 11.4484084447225
572 11.1421152750651
573 11.1406723658244
574 11.0462945302327
575 10.9755048751831
576 10.8437531789144
577 10.7807900110881
578 10.5630610783895
579 10.3815253575643
580 10.3279218673706
581 10.3787310918172
582 10.5737721125285
583 10.9023898442586
584 11.0432284673055
585 11.4016860326131
586 11.8158439000448
587 12.6417878468831
588 13.2347774505615
589 12.6336638132731
590 11.2933330535889
591 10.5735769271851
592 10.7139132817586
593 10.8686051368713
594 10.6277370452881
595 10.2225289344788
596 10.1385037104289
597 10.420389175415
598 10.8423382441203
599 11.5262854894002
600 11.8505601882935
601 11.8350435892741
602 11.4390589396159
603 11.1412947972616
604 10.913680712382
605 10.4885733922323
606 10.1810642878215
607 10.1177911758423
608 10.1319360733032
609 10.2568122545878
610 10.3543691635132
611 10.5073448816935
612 10.6739590962728
613 10.913415590922
614 11.3091564178467
615 12.1234118143717
616 12.5250024795532
617 12.3531999588013
618 11.8108800252279
619 11.260051091512
620 10.7166624069214
621 10.5704663594564
622 10.6384048461914
623 10.6574047406514
624 10.4404323895772
625 10.1648108164469
626 10.0267766316732
627 10.0748047828674
628 10.292799949646
629 10.7447551091512
630 11.295778910319
631 11.9446217219035
632 11.931173324585
633 11.9391307830811
634 11.2791086832682
635 10.6041863759359
636 10.2994607289632
637 10.1836673418681
638 10.1386499404907
639 10.153866926829
640 10.1535116831462
641 10.2282632191976
642 10.2832074165344
643 10.2224206924438
644 10.2265858650208
645 10.3474518458048
646 10.755412419637
647 11.4284925460815
648 12.0726620356242
649 12.5596691767375
650 11.7531264623006
651 11.4005489349365
652 11.0875558853149
653 10.7132275899251
654 10.4771560033162
655 10.3903690973918
656 10.3158340454102
657 10.1524106661479
658 10.0613686243693
659 10.0421215693156
660 10.0731023152669
661 10.2349278132121
662 10.5951992670695
663 11.1256472269694
664 11.3668543497721
665 11.7882566452026
666 11.9100977579753
667 11.4612719217936
668 10.8124707539876
669 10.2957283655802
670 9.99824476242065
671 9.98135153452555
672 10.0865782101949
673 10.0857601165771
674 10.033732732137
675 10.046701113383
676 10.0990368525187
677 10.2834943135579
678 10.6795225143433
679 11.2760187784831
680 11.5129302342733
681 11.9881639480591
682 11.999875386556
683 11.1615826288859
684 10.4223925272624
685 10.2952852249146
686 10.2479499181112
687 10.1829005877177
688 10.1074331601461
689 10.1320530573527
690 10.2763175964355
691 10.650912920634
692 10.5941953659058
693 10.5530869166056
694 10.6870794296265
695 10.9115513165792
696 11.116117477417
697 11.048254330953
698 10.8408377965291
699 10.5951999028524
700 10.1399408976237
701 9.85081402460734
702 9.72017018000285
703 9.72985617319743
704 9.75716908772787
705 9.78834168116252
706 9.85562165578206
707 10.0873638788859
708 10.4775724411011
709 11.0322179794312
710 11.607271194458
711 11.9437306722005
712 11.9798994064331
713 11.8102722167969
714 11.1011969248454
715 10.4757582346598
716 10.3497638702393
717 10.4837220509847
718 10.4093055725098
719 10.315967241923
720 10.1296895345052
721 10.1285489400228
722 10.0959749221802
723 10.2452545166016
724 10.3768239021301
725 10.5103011131287
726 10.48331006368
727 10.3522043228149
728 10.3765999476115
729 10.5392011006673
730 10.594978650411
731 10.3932151794434
732 10.1598221460978
733 9.94947131474813
734 9.9337674776713
735 9.97375647226969
736 9.98463694254557
737 9.95031070709229
738 9.92147397994995
739 9.92759164174398
740 9.96851523717245
741 10.0520000457764
742 10.1949520111084
743 10.3554770151774
744 10.4759383201599
745 10.3343175252279
746 10.1502944628398
747 9.91194629669189
748 9.85406764348348
749 9.78597704569499
750 9.72089989980062
751 9.6688969930013
752 9.67590236663818
753 9.70013173421224
754 9.80610958735148
755 9.90156602859497
756 10.0511171023051
757 10.1417053540548
758 10.224062760671
759 10.1018582979838
760 9.93189255396525
761 9.69223674138387
762 9.56321907043457
763 9.46994988123576
764 9.47302373250326
765 9.48911603291829
766 9.57459084192912
767 9.65491692225138
768 9.82993141810099
769 10.0545775095622
770 10.4316349029541
771 10.8420257568359
772 11.0228026707967
773 10.9785757064819
774 10.9927082061768
775 11.0026556650798
776 11.0594088236491
777 11.5375889142354
778 12.2157694498698
779 11.7908263206482
780 10.3660233815511
781 10.1580540339152
782 10.6253188451131
783 10.4347631136576
784 10.1279045740763
785 9.81224473317464
786 9.80556456247965
787 9.8398593266805
788 10.0047801335653
789 10.3901241620382
790 10.6917848587036
791 11.0223817825317
792 10.8843406041463
793 10.2858762741089
794 9.80816062291463
795 9.62781000137329
796 9.55790026982625
797 9.51754633585612
798 9.52566067377726
799 9.47481807072957
800 9.46530373891195
801 9.50342480341593
802 9.59802436828613
803 9.78852876027425
804 10.0369097391764
805 10.27903175354
806 10.3235413233439
807 10.289444287618
808 10.1287097930908
809 9.90926027297974
810 9.70169798533122
811 9.56010214487712
812 9.51843070983887
813 9.53569968541463
814 9.61916955312093
815 9.72088130315145
816 9.88893175125122
817 10.0923659006755
818 10.3936023712158
819 10.8457686106364
820 11.8424695332845
821 13.0369189580282
822 13.5838680267334
823 13.4412336349487
824 12.1949030558268
825 11.3436012268066
826 11.6534026463827
827 11.6264147758484
828 10.8398914337158
829 10.3746178944906
830 10.8296092351278
831 10.7224802970886
832 10.0937280654907
833 9.72685861587524
834 9.61613972981771
835 9.54479551315308
836 9.44015407562256
837 9.30908632278442
838 9.31784598032633
839 9.29579242070516
840 9.28788503011068
841 9.28236834208171
842 9.3080792427063
843 9.38389190038045
844 9.45255184173584
845 9.57229868570963
846 9.6248099009196
847 9.70605246225993
848 9.72329139709473
849 9.80045509338379
850 9.84110244115194
851 9.81544001897176
852 9.67526769638062
853 9.51700894037882
854 9.3798885345459
855 9.36117855707804
856 9.36778624852498
857 9.45608965555827
858 9.55583222707113
859 9.73975721995036
860 9.98502826690674
861 10.4469216664632
862 11.1813166936239
863 12.1896003087362
864 13.1924053827922
865 12.4298375447591
866 10.8338034947713
867 10.0952682495117
868 10.3319239616394
869 10.5744651158651
870 10.5032720565796
871 10.3074746131897
872 10.0659898122152
873 10.0227282842
874 10.0202946662903
875 10.2445085843404
876 10.3414169947306
877 10.051487604777
878 9.82879177729289
879 9.82168515523275
880 9.91757043202718
881 9.99191490809123
882 9.93275133768717
883 9.63570721944173
884 9.41568565368652
885 9.43021949132284
886 9.42739248275757
887 9.38880491256714
888 9.3241974512736
889 9.28526512781779
890 9.30713860193888
891 9.34128411610921
892 9.40837017695109
893 9.51163800557454
894 9.64547348022461
895 9.80756600697835
896 10.0013122558594
897 10.0337521235148
898 9.99880758921305
899 9.80575116475423
900 9.7173433303833
901 9.62114874521891
902 9.52827739715576
903 9.42090733846029
904 9.37532901763916
905 9.36242961883545
906 9.4441343943278
907 9.52251100540161
908 9.6254022916158
909 9.75342162450155
910 9.94747765858968
911 10.2060724894206
912 10.2719411849976
913 10.2582043011983
914 10.1196775436401
915 10.0236248970032
916 9.88381052017212
917 9.77389637629191
918 9.74479595820109
919 9.77667156855265
920 9.76728852589925
921 9.73917531967163
922 9.63031514485677
923 9.5849126180013
924 9.56513547897339
925 9.56307458877563
926 9.57387288411458
927 9.54944483439128
928 9.52414623896281
929 9.55467446645101
930 9.60664542516073
931 9.6823468208313
932 9.74862988789876
933 9.79296716054281
934 9.74992497762044
935 9.70891745885213
936 9.58866453170776
937 9.55452092488607
938 9.57297197977702
939 9.70251846313477
940 9.89322566986084
941 10.0134881337484
942 10.0683018366496
943 9.8982048034668
944 9.76350275675456
945 9.62942091623942
946 9.66412814458211
947 9.81201553344727
948 10.1588220596313
949 10.7848081588745
950 11.7633120218913
951 12.7089255650838
952 12.4527568817139
953 11.4540667533875
954 10.2253403663635
955 10.2947158813477
956 10.2865022023519
957 10.0935718218486
958 10.0461797714233
959 10.0196730295817
960 9.93732579549154
961 9.84701951344808
962 9.84428882598877
963 9.87210671106974
964 10.0828696886698
965 10.4228137334188
966 10.7151258786519
967 10.9188466072083
968 10.5413737297058
969 9.82218217849731
970 9.47855536142985
971 9.38581927617391
972 9.35112984975179
973 9.30144278208415
974 9.27674611409505
975 9.34099197387695
976 9.46712875366211
977 9.48048448562622
978 9.50959491729736
979 9.53307294845581
980 9.65740474065145
981 9.83502356211344
982 10.0102033615112
983 10.2400000890096
984 10.5330794652303
985 10.8097887039185
986 10.6188720067342
987 10.2648591995239
988 9.84182739257812
989 9.55927181243896
990 9.4099858601888
991 9.33380953470866
992 9.35570557912191
993 9.36477470397949
994 9.43362840016683
995 9.40343936284383
996 9.42726310094198
997 9.41980600357056
998 9.45390510559082
999 9.41623163223267
1000 9.44648949305216
1001 9.48526191711426
1002 9.63308080037435
1003 9.87830670674642
1004 10.1749717394511
1005 10.5185848871867
1006 10.668031056722
1007 10.5657725334167
1008 10.0395302772522
1009 9.64162842432658
1010 9.35569667816162
1011 9.27088483174642
1012 9.27278598149618
1013 9.2828885714213
1014 9.33427254358927
1015 9.33837938308716
1016 9.42209053039551
1017 9.45334943135579
1018 9.52431217829386
1019 9.48784160614014
1020 9.49639876683553
1021 9.43748537699381
1022 9.53197463353475
1023 9.63223266601562
1024 9.87281179428101
1025 10.1030944188436
1026 10.1954536437988
1027 10.1240844726562
1028 9.84146006902059
1029 9.69740549723307
1030 9.56650813420614
1031 9.51145299275716
1032 9.41601149241129
1033 9.32487948735555
1034 9.29907973607381
1035 9.31423981984456
1036 9.30872774124146
1037 9.32353782653809
1038 9.3874659538269
1039 9.50936730702718
1040 9.69630177815755
1041 9.72705777486165
1042 9.71075407663981
1043 9.60405158996582
1044 9.60436582565308
1045 9.61581643422445
1046 9.71231810251872
1047 9.75300455093384
1048 9.79741986592611
1049 9.76649284362793
1050 9.6434539159139
1051 9.50432872772217
1052 9.35955333709717
1053 9.34789594014486
1054 9.39354165395101
1055 9.57514047622681
1056 9.8428438504537
1057 10.134904384613
1058 10.1459137598673
1059 10.104425907135
1060 9.74035580952962
1061 9.52079089482625
1062 9.32070096333822
1063 9.30059178670247
1064 9.46048307418823
1065 9.62954648335775
1066 9.91245206197103
1067 10.1147990226746
1068 10.3410936991374
1069 10.4026959737142
1070 10.4427386919657
1071 9.91975386937459
1072 9.45811684926351
1073 9.18516413370768
1074 9.11818917592367
1075 9.11063607533773
1076 9.09639644622803
1077 9.16150045394897
1078 9.17879772186279
1079 9.2840690612793
1080 9.40200773874919
1081 9.54053370157878
1082 9.70646651585897
1083 9.90287526448568
1084 10.006156762441
1085 10.2910785675049
1086 10.8078393936157
1087 11.4111928939819
1088 11.807715733846
1089 11.1067717870077
1090 10.5540558497111
1091 9.90710671742757
1092 9.66722377141317
1093 9.74666245778402
1094 9.7915096282959
1095 9.73638153076172
1096 9.61339314778646
1097 9.66015466054281
1098 9.66986211140951
1099 9.78311475118001
1100 9.90584595998128
1101 9.90246836344401
1102 9.75582949320475
1103 9.79010057449341
1104 9.89092763264974
1105 9.90277703603109
1106 9.82365004221598
1107 9.69876861572266
1108 9.75454998016357
1109 9.80776166915894
1110 9.74488719304403
1111 9.40292978286743
1112 9.23567708333333
1113 9.24312019348145
1114 9.22207276026408
1115 9.22306378682454
1116 9.19137636820475
1117 9.23443190256754
1118 9.24577442804972
1119 9.24905935923258
1120 9.3017144203186
1121 9.39095544815063
1122 9.54546387990316
1123 9.67667881647746
1124 9.85240793228149
1125 9.88114992777506
1126 9.77191638946533
1127 9.53327719370524
1128 9.34863901138306
1129 9.18719959259033
1130 9.11032772064209
1131 9.06397072474162
1132 9.05081653594971
1133 9.07593250274658
1134 9.10578505198161
1135 9.1793114344279
1136 9.26471598943075
1137 9.41221555074056
1138 9.47237014770508
1139 9.58025534947713
1140 9.55815553665161
1141 9.67693980534871
1142 9.7559757232666
1143 9.84528493881226
1144 9.79458300272624
1145 9.64806572596232
1146 9.47046931584676
1147 9.35078128178914
1148 9.29692840576172
1149 9.26669375101725
1150 9.24901453653971
1151 9.31019878387451
1152 9.39547872543335
1153 9.5905917485555
1154 9.80936908721924
1155 10.0110759735107
1156 10.1566580136617
1157 9.98227039972941
1158 9.66541957855225
1159 9.31198867162069
1160 9.12340943018595
1161 9.05454158782959
1162 9.0498997370402
1163 9.14406887690226
1164 9.2136058807373
1165 9.33790175120036
1166 9.34846973419189
1167 9.45853360493978
1168 9.53364181518555
1169 9.73434066772461
1170 9.66670481363932
1171 9.64096895853678
1172 9.62864764531454
1173 9.71331898371379
1174 9.96989583969116
1175 10.4134097099304
1176 11.0655692418416
1177 11.4696051279704
1178 11.2707767486572
1179 10.4510545730591
1180 10.0853713353475
1181 9.74350786209106
1182 9.78869787851969
1183 9.93719625473022
1184 9.83145189285278
1185 9.75663820902506
1186 9.85333188374837
1187 9.81361436843872
1188 9.73057206471761
1189 9.83632246653239
1190 9.89644320805868
1191 9.77921438217163
1192 9.74199231465658
1193 9.72566302617391
1194 9.80175240834554
1195 9.80249849955241
1196 9.93872769673665
1197 9.7942697207133
1198 9.59361521402995
1199 9.39621814092
1200 9.33457136154175
1201 9.24304072062174
1202 9.15114291508993
1203 9.09492365519206
1204 9.11391909917196
1205 9.16559362411499
1206 9.2240948677063
1207 9.24845949808756
1208 9.33186896642049
1209 9.43469874064128
1210 9.60562896728516
1211 9.68873945871989
1212 9.78074200948079
1213 9.6750537554423
1214 9.53550163904826
1215 9.29014333089193
1216 9.16960446039836
1217 9.06668122609456
1218 9.03258673350016
1219 9.019988377889
1220 8.97924598058065
1221 8.98271799087524
1222 9.00681591033936
1223 9.04330444335938
1224 9.09986193974813
1225 9.20231469472249
1226 9.29529603322347
1227 9.44320694605509
1228 9.51616795857747
1229 9.61712042490641
1230 9.63434219360352
1231 9.61655267079671
1232 9.62167739868164
1233 9.60552310943604
1234 9.61888694763184
1235 9.54080963134766
1236 9.47035090128581
1237 9.41910632451375
1238 9.38734070460002
1239 9.22047392527262
1240 9.12289603551229
1241 9.06141312917074
1242 9.09737841288249
1243 9.19248072306315
1244 9.25775750478109
1245 9.25243504842122
1246 9.15697924296061
1247 9.09845574696859
1248 9.07649183273315
1249 9.02636400858561
1250 8.96601390838623
1251 8.91007089614868
1252 8.89021682739258
1253 8.8888619740804
1254 8.92924865086873
1255 8.99854421615601
1256 9.1120719909668
1257 9.15253098805746
1258 9.15757290522257
1259 9.11824814478556
1260 9.13879044850667
1261 9.23775641123454
1262 9.38564904530843
1263 9.54884672164917
1264 9.62022972106934
1265 9.73110580444336
1266 9.73322327931722
1267 9.78180074691772
1268 9.70288562774658
1269 9.60661729176839
1270 9.44845787684123
1271 9.45676898956299
1272 9.56311798095703
1273 9.77562506993612
1274 10.0218243598938
1275 10.3741979598999
1276 10.2358964284261
1277 10.130989074707
1278 9.83200788497925
1279 9.51930522918701
1280 9.27712869644165
1281 9.11780548095703
1282 8.99921814600627
1283 8.96946779886881
1284 9.0758752822876
1285 9.1810843149821
1286 9.42233769098918
1287 9.37999089558919
1288 9.32224750518799
1289 9.22550741831462
1290 9.24434343973796
1291 9.38923327128092
1292 9.57404152552287
1293 9.77729431788127
1294 9.92630608876546
1295 10.1760473251343
1296 10.1863590876261
1297 10.3278144200643
1298 9.96746921539307
1299 9.46701288223267
1300 9.08034944534302
1301 9.03865083058675
1302 9.06083456675212
1303 9.19139099121094
1304 9.2402712504069
1305 9.34350331624349
1306 9.53219350179037
1307 9.67771943410238
1308 9.87138573328654
1309 10.1248450279236
1310 10.0442056655884
1311 9.85705995559692
1312 9.48318481445312
1313 9.45800399780273
1314 9.50077931086222
1315 9.50427802403768
1316 9.38406324386597
1317 9.44889370600382
1318 9.59852774937948
1319 9.7269868850708
1320 9.64824835459391
1321 9.35022131601969
1322 9.40880250930786
1323 9.58187166849772
1324 9.68956232070923
1325 9.95525328318278
1326 10.040011882782
1327 9.68515125910441
1328 9.42582766215006
1329 9.24866326649984
1330 9.14936828613281
1331 9.2655709584554
1332 9.45217847824097
1333 9.52792930603027
1334 9.39835278193156
1335 9.35457674662272
1336 9.31461890538534
1337 9.38562679290771
1338 9.52972793579102
1339 9.44759607315063
1340 9.3921545346578
1341 9.37974103291829
1342 9.4524720509847
1343 9.37630208333333
1344 9.34691079457601
1345 9.34300867716471
1346 9.29664691289266
1347 9.31834348042806
1348 9.42367394765218
1349 9.55795764923096
1350 9.67148001988729
1351 9.87366930643717
1352 9.63562901814779
1353 9.56332111358643
1354 9.62673695882162
1355 9.77134354909261
1356 9.85773309071859
1357 9.91438961029053
1358 9.75903415679932
1359 9.64818636576335
1360 9.42722050348918
1361 9.35332536697388
1362 9.24702962239583
1363 9.26354153951009
1364 9.27395804723104
1365 9.34898042678833
1366 9.31191698710124
1367 9.30571778615316
1368 9.29107173283895
1369 9.40275573730469
1370 9.50743405024211
1371 9.68756373723348
1372 9.82334423065186
1373 9.88945945103963
1374 9.79810746510824
1375 9.61386855443319
1376 9.33994960784912
1377 9.22738838195801
1378 9.13017845153809
1379 9.13816388448079
1380 9.21844005584717
1381 9.20867125193278
1382 9.25580596923828
1383 9.28023179372152
1384 9.35291353861491
1385 9.41547616322835
1386 9.5067933400472
1387 9.49032354354858
1388 9.50396839777629
1389 9.36701949437459
1390 9.32091840108236
1391 9.35828495025635
1392 9.50138441721598
1393 9.83192539215088
1394 10.2784667015076
1395 10.5436396598816
1396 10.4466571807861
1397 10.220311164856
1398 9.80563958485921
1399 9.597412109375
1400 9.30280462900797
1401 9.1708083152771
1402 9.15142726898193
1403 9.14627599716187
1404 9.17577330271403
1405 9.23964500427246
1406 9.27091058095296
1407 9.22021309534709
1408 9.23331069946289
1409 9.27674166361491
1410 9.38779067993164
1411 9.4642333984375
1412 9.57354958852132
1413 9.62922096252441
1414 9.67163403828939
1415 9.79666519165039
1416 9.90869585673014
1417 10.0608649253845
1418 9.80316193898519
1419 9.73253281911214
1420 9.49361451466878
1421 9.43173042933146
1422 9.37283054987589
1423 9.17948849995931
1424 9.05204900105794
1425 9.04972155888875
1426 9.08485317230225
1427 9.20215257008871
1428 9.32761144638062
1429 9.39007727305094
1430 9.46801567077637
1431 9.41378672917684
1432 9.52121384938558
1433 9.78330294291178
1434 10.1291913986206
1435 10.4008007049561
1436 10.1224435170492
1437 10.0538206100464
1438 9.90937042236328
1439 9.46319993336996
1440 9.27069727579753
1441 9.20561440785726
1442 9.20946582158407
1443 9.29124863942464
1444 9.34799559911092
1445 9.4247883160909
1446 9.59733231862386
1447 9.63115167617798
1448 9.58184846242269
1449 9.58743699391683
1450 9.60420370101929
1451 9.78042014439901
1452 9.94890149434408
1453 10.0643982887268
1454 9.96742916107178
1455 9.89404678344727
1456 9.53286902109782
1457 9.27289930979411
1458 9.21640237172445
1459 9.18411159515381
1460 9.28725020090739
1461 9.36522324879964
1462 9.38727728525797
1463 9.39890718460083
1464 9.45476977030436
1465 9.4299160639445
1466 9.41887633005778
1467 9.45101340611776
1468 9.55073658625285
1469 9.77989562352498
1470 9.92000230153402
1471 9.85872952143351
1472 9.71176862716675
1473 9.68781677881877
1474 9.36032867431641
1475 9.23156277338664
1476 9.17216396331787
1477 9.11093934377035
1478 9.16849374771118
1479 9.18663231531779
1480 9.17069594065348
1481 9.24093246459961
1482 9.39387178421021
1483 9.41764068603516
1484 9.45931800206502
1485 9.55255540211996
1486 9.63857158025106
1487 9.73374223709106
1488 9.80908075968424
1489 9.73747062683105
1490 9.61483796437581
1491 9.56699371337891
1492 9.29186820983887
1493 9.06926123301188
1494 9.07290633519491
1495 9.08452256520589
1496 9.07738653818766
1497 9.12443749109904
1498 9.17497316996256
1499 9.20334386825562
1500 9.28240219751994
1501 9.33660538991292
1502 9.37258847554525
1503 9.41278998057047
1504 9.49816735585531
1505 9.67074155807495
1506 9.88098494211833
1507 10.0795594851176
1508 10.0301311810811
1509 9.93987035751343
1510 9.58522081375122
1511 9.25459973017375
1512 9.03316148122152
1513 9.01378695170085
1514 9.05057938893636
1515 9.10411357879639
1516 9.13436079025269
1517 9.19571050008138
1518 9.32729848225912
1519 9.44978380203247
1520 9.47914981842041
1521 9.44044733047485
1522 9.44688940048218
1523 9.51842530568441
1524 9.64113807678223
1525 9.76249186197917
1526 9.75209967295329
1527 9.72576030095418
1528 9.57642904917399
1529 9.51757971445719
1530 9.17677688598633
1531 8.98112853368123
1532 8.9340828259786
1533 8.93801641464233
1534 8.97864659627279
1535 9.04892762502035
1536 9.10373767217
1537 9.1937902768453
1538 9.31059869130453
1539 9.38416639963786
1540 9.40474764506022
1541 9.38906558354696
1542 9.3903021812439
1543 9.48377339045207
1544 9.61814721425374
1545 9.74499162038167
1546 9.69491116205851
1547 9.65856122970581
1548 9.54426495234171
1549 9.4646209081014
1550 9.12217330932617
1551 8.95439481735229
1552 8.933571656545
1553 8.98134740193685
1554 9.04132095972697
1555 9.09868590037028
1556 9.1472110748291
1557 9.21080128351847
1558 9.31445487340291
1559 9.42500734329224
1560 9.52200396855672
1561 9.6084246635437
1562 9.74821949005127
1563 9.88167858123779
1564 9.83912452061971
1565 9.67305246988932
1566 9.45215431849162
1567 9.3985071182251
1568 9.29942401250204
1569 9.22046454747518
1570 9.06798871358236
1571 9.01070594787598
1572 9.00736141204834
1573 9.09704033533732
1574 9.19474522272746
1575 9.26312716801961
1576 9.29898484547933
1577 9.24961312611898
1578 9.27990102767944
1579 9.31254784266154
1580 9.39210955301921
1581 9.47494029998779
1582 9.62244574228922
1583 9.86154905954997
1584 10.1204821268717
1585 10.2228609720866
1586 9.77685578664144
1587 9.39260705312093
1588 9.08571306864421
1589 8.93797969818115
1590 8.88795598347982
1591 8.96358029047648
1592 9.01266876856486
1593 9.01382033030192
1594 9.10239839553833
1595 9.30687634150187
1596 9.55275472005208
1597 9.74546893437704
1598 9.76993131637573
1599 9.73721297581991
1600 9.40310621261597
1601 9.23616790771484
1602 9.35615317026774
1603 9.68235317866007
1604 9.82692750295003
1605 9.72633091608683
1606 9.30552832285563
1607 9.1366491317749
1608 9.15388711293538
1609 9.19067319234212
1610 9.16529989242554
1611 9.07678461074829
1612 9.02970170974731
1613 8.99763154983521
1614 8.94258117675781
1615 8.92359256744385
1616 8.88764603932699
1617 8.91281000773112
1618 8.91513554255168
1619 8.95465183258057
1620 8.94636074701945
1621 8.93137756983439
1622 8.89508676528931
1623 8.92031669616699
1624 8.94295326868693
1625 9.03128544489543
1626 9.09503253300985
1627 9.2357177734375
1628 9.35036627451579
1629 9.43019390106201
1630 9.29756164550781
1631 9.20769818623861
1632 9.05124632517497
1633 8.97730286916097
1634 8.89430459340413
1635 8.87912480036418
1636 8.86451466878255
1637 8.85592873891195
1638 8.84350744883219
1639 8.83267752329508
1640 8.82284927368164
1641 8.86267948150635
1642 8.927521387736
1643 9.04287560780843
1644 9.17108392715454
1645 9.34774287541707
1646 9.4088986714681
1647 9.5673942565918
1648 9.76820580164591
1649 9.90065542856852
1650 9.90456120173136
1651 9.60134569803874
1652 9.26651271184286
1653 9.06394863128662
1654 9.04278008143107
1655 9.0816650390625
1656 9.12876653671265
1657 9.17576138178507
1658 9.21118386586507
1659 9.28776677449544
1660 10.6139653523763
1661 12.8157780965169
1662 11.3956613540649
1663 11.6733350753784
1664 12.1074962615967
1665 12.0959850947062
1666 11.8590437571208
1667 11.6758311589559
1668 11.0595719019572
1669 11.3952328364054
1670 12.3522439002991
1671 12.5089279810588
1672 11.3755084673564
1673 11.0018393198649
1674 11.17032178243
1675 10.6476833025614
1676 11.1856517791748
1677 12.5799999237061
1678 12.1193563143412
1679 11.2842334111532
1680 11.0634651184082
1681 10.5601607958476
1682 10.318025747935
1683 10.3341811498006
1684 10.2940810521444
1685 10.3443004290263
1686 10.0178596178691
1687 9.86782550811768
1688 9.85481929779053
1689 9.66803820927938
1690 9.28965600331624
1691 9.26678117116292
1692 9.38112004597982
1693 9.41400130589803
1694 9.41395950317383
1695 9.30698235829671
1696 9.21055348714193
1697 9.18297497431437
1698 9.19397083918254
1699 9.19470151265462
1700 9.2101944287618
1701 9.20453135172526
1702 9.20169258117676
1703 9.18597094217936
1704 9.16705958048503
1705 9.14084800084432
1706 9.12775150934855
1707 9.13318204879761
1708 9.14412514368693
1709 9.17114543914795
1710 9.20502312978109
1711 9.23588387171427
1712 9.25557867685954
1713 9.25643348693848
1714 9.24648729960124
1715 9.23969237009684
1716 9.23130432764689
1717 9.2318229675293
1718 9.23341369628906
1719 9.23582681020101
1720 9.23934523264567
1721 9.24282964070638
1722 9.2363494237264
1723 9.22460746765137
1724 9.21043745676676
1725 9.20255740483602
1726 9.19690529505412
1727 9.20116265614828
1728 9.20910199483236
1729 9.21729548772176
1730 9.22436745961507
1731 9.22486670811971
1732 9.21542803446452
1733 9.2013053894043
1734 9.18777704238892
1735 9.17766698201497
1736 9.1731497446696
1737 9.17314163843791
1738 9.1786363919576
1739 9.17857217788696
1740 9.18125581741333
1741 9.17822869618734
1742 9.17674954732259
1743 9.17411470413208
1744 9.17423057556152
1745 9.17141183217367
1746 9.1735676129659
1747 9.17862415313721
1748 9.18663152058919
1749 9.19486490885417
1750 9.20497751235962
1751 9.21115763982137
1752 9.21766503651937
1753 9.22335052490234
1754 9.23313124974569
1755 9.24402046203613
1756 9.26320918401082
1757 9.28420766194662
1758 9.31303866704305
1759 9.34587144851685
1760 9.38234376907349
1761 9.42200978597005
1762 9.46554597218831
1763 9.53778123855591
1764 9.59825690587362
1765 9.69921938578288
1766 9.7021427154541
1767 9.73171234130859
1768 9.60926723480225
1769 9.46516640981038
1770 9.3243883450826
1771 9.20527219772339
1772 9.14719359079997
1773 9.12952979405721
1774 9.12785641352336
1775 9.15394147237142
1776 9.19858423868815
1777 9.25299803415934
1778 9.31047058105469
1779 9.41024748484294
1780 9.52165540059408
1781 9.70165713628133
1782 9.85912847518921
1783 10.04731909434
1784 10.0528512001038
1785 10.0315558115641
1786 9.8752924601237
1787 9.77129872639974
1788 9.63913186391195
1789 9.48754930496216
1790 9.33426872889201
1791 9.24310525258382
1792 9.17758973439535
1793 9.16908359527588
1794 9.1780579884847
1795 9.19375610351562
1796 9.22535626093546
1797 9.33535162607829
1798 9.48171250025431
1799 9.65134922663371
1800 9.75492032368978
1801 9.89990520477295
1802 9.96940501530965
1803 10.0521376927694
1804 9.93582741419474
1805 9.78643051783244
1806 9.63146162033081
1807 9.58109903335571
1808 9.48628584543864
1809 9.34862279891968
1810 9.22745609283447
1811 9.18897724151611
1812 9.22414843241374
1813 9.34332863489787
1814 9.44751803080241
1815 9.57251691818237
1816 9.61252498626709
1817 9.67457977930705
1818 9.74277114868164
1819 9.86120557785034
1820 9.8336607615153
1821 9.78465366363525
1822 9.64268747965495
1823 9.6145076751709
1824 9.64018201828003
1825 9.78683884938558
1826 9.91090297698975
1827 9.95111242930094
1828 9.92929712931315
1829 9.93659083048503
1830 9.7286966641744
1831 9.53552039464315
1832 9.43781391779582
1833 9.44384606679281
1834 9.43951908747355
1835 9.4652369817098
1836 9.47208293279012
1837 9.45671129226685
1838 9.33778206507365
1839 9.25956439971924
1840 9.22039190928141
1841 9.20644474029541
1842 9.13253943125407
1843 9.07255458831787
1844 9.07999356587728
1845 9.12464014689128
1846 9.20543829600016
1847 9.22952365875244
1848 9.27352889378866
1849 9.42229032516479
1850 9.65026378631592
1851 9.62262344360352
1852 9.35020844141642
1853 9.17286745707194
1854 9.12259292602539
1855 9.12397543589274
1856 9.06230481465658
1857 9.03366533915202
1858 9.04808457692464
1859 9.09531545639038
1860 9.14619032541911
1861 9.20439465840658
1862 9.23574876785278
1863 9.30018552144369
1864 9.34900283813477
1865 9.43432299296061
1866 9.49602063496908
1867 9.60325336456299
1868 9.61579672495524
1869 9.66897265116374
1870 9.61010583241781
1871 9.54512453079224
1872 9.43795982996623
1873 9.36561266581217
1874 9.30183744430542
1875 9.2965202331543
1876 9.28226772944132
1877 9.30891879399618
1878 9.34994188944499
1879 9.40646012624105
1880 9.43427276611328
1881 9.44842656453451
1882 9.44762007395426
1883 9.48759746551514
1884 9.53875700632731
1885 9.57393503189087
1886 9.53171745936076
1887 9.45110956827799
1888 9.35943110783895
1889 9.3124262491862
1890 9.24054177602132
1891 9.21142228444417
1892 9.16135342915853
1893 9.12793032328288
1894 9.08077557881673
1895 9.05353387196859
1896 9.02373250325521
1897 9.03571287790934
1898 9.08273379007975
1899 9.19985548655192
1900 9.37445608774821
1901 9.53363498051961
1902 9.55019585291545
1903 9.49000946680705
1904 9.37687842051188
1905 9.30119466781616
1906 9.2803963025411
1907 9.26647774378459
1908 9.30479876200358
1909 9.33831596374512
1910 9.38541142145793
1911 9.33395020167033
1912 9.26419051488241
1913 9.25170310338338
1914 9.18616819381714
1915 9.12164831161499
1916 9.04103549321493
1917 9.00346247355143
1918 9.00006230672201
1919 9.02972571055094
1920 9.08397738138835
1921 9.17033863067627
1922 9.26358811060588
1923 9.38464959462484
1924 9.46980174382528
1925 9.52699247996012
1926 9.50410048166911
1927 9.44767506917318
1928 9.34695943196615
1929 9.26058975855509
1930 9.18742211659749
1931 9.12315400441488
1932 9.07448768615723
1933 9.04507414499919
1934 9.03787469863892
1935 9.05997737248739
1936 9.10607862472534
1937 9.18435462315877
1938 9.26163864135742
1939 9.33467785517375
1940 9.35893090565999
1941 9.36704651514689
1942 9.34972635904948
1943 9.36125342051188
1944 9.38111718495687
1945 9.46844053268433
1946 9.57031965255737
1947 9.68849452336629
1948 9.75117683410645
1949 9.82683038711548
1950 9.7895892461141
1951 9.75482145945231
1952 9.47013600667318
1953 9.38029511769613
1954 9.28245449066162
1955 9.24034786224365
1956 9.29899613062541
1957 9.40658378601074
1958 9.52419551213582
1959 9.63911120096842
1960 9.7699613571167
1961 9.96537844340006
1962 9.81510893503825
1963 9.58156855901082
1964 9.47104056676229
1965 9.46805413564046
1966 9.46849219004313
1967 9.52582629521688
1968 9.44068463643392
1969 9.35880406697591
1970 9.30051136016846
1971 9.3002192179362
1972 9.26329056421916
1973 9.23098532358805
1974 9.16820462544759
1975 9.09986193974813
1976 9.07971318562826
1977 9.12226152420044
1978 9.19332456588745
1979 9.32802788416545
1980 9.45246458053589
1981 9.47640117009481
1982 9.40071678161621
1983 9.32202990849813
1984 9.23115253448486
1985 9.25006437301636
1986 9.18530591328939
1987 9.09716113408407
1988 9.01179122924805
1989 8.97521305084229
1990 9.02153587341309
1991 9.0385487874349
1992 9.02021996180216
1993 9.03642082214355
1994 9.0341321627299
1995 9.06836303075155
1996 9.10085217158
1997 9.11838404337565
1998 9.1640051205953
1999 9.24579334259033
2000 9.28511428833008
};
\end{axis}

\end{tikzpicture}

%% file: nomseplot.tex
\begin{tikzpicture}[scale = 0.55]

\definecolor{darkgray176}{RGB}{176,176,176}

\begin{axis}[
tick align=outside,
tick pos=left,
x grid style={darkgray176},
xlabel={Number of Epochs},
xmajorgrids,
xmin=-98.95, xmax=2099.95,
xtick style={color=black},
y grid style={darkgray176},
ylabel={Training Loss},
ymajorgrids,
ymin=0.0199371345962087, ymax=0.320447697676718,
ytick style={color=black}
]
\addplot [semithick, blue]
table {%
1 0.306788126627604
2 0.263048708438873
3 0.259596019983292
4 0.257297496000926
5 0.25750199953715
6 0.251455614964167
7 0.245376487572988
8 0.241204296549161
9 0.238216256101926
10 0.233180657029152
11 0.227361932396889
12 0.222653026382128
13 0.217525924245516
14 0.213369910915693
15 0.208656132221222
16 0.207422067721685
17 0.204597850640615
18 0.202832822998365
19 0.19214990735054
20 0.182210067907969
21 0.171324690183004
22 0.161933213472366
23 0.158396939436595
24 0.154411405324936
25 0.148104310035706
26 0.142990122238795
27 0.13900588452816
28 0.134463255604108
29 0.131197080016136
30 0.127901824812094
31 0.126737512648106
32 0.122664590676626
33 0.121272166570028
34 0.118689519663652
35 0.115436665713787
36 0.110777743160725
37 0.105340922872225
38 0.0996403594811757
39 0.0933127452929815
40 0.0899577463666598
41 0.0887691701451937
42 0.0862295205394427
43 0.0847393274307251
44 0.0879352788130442
45 0.0863927404085795
46 0.0825953682263692
47 0.0838284865021706
48 0.0772645125786463
49 0.0786551907658577
50 0.0760352884729703
51 0.0760388274987539
52 0.0733521257837613
53 0.0729214623570442
54 0.072715329627196
55 0.0715507417917252
56 0.0709823022286097
57 0.0699988752603531
58 0.0698745523889859
59 0.0686913753549258
60 0.068442739546299
61 0.0676156828800837
62 0.0673519521951675
63 0.0669645021359126
64 0.0666492655873299
65 0.066593162715435
66 0.0667963027954102
67 0.0661676004528999
68 0.0665479078888893
69 0.0675983677307765
70 0.068669522802035
71 0.0730483333269755
72 0.0817227686444918
73 0.0791970392068227
74 0.0813059185942014
75 0.0819476346174876
76 0.0757390707731247
77 0.0753593395153681
78 0.0747742205858231
79 0.0716281235218048
80 0.0721280574798584
81 0.0718296443422635
82 0.0700046867132187
83 0.0686612154046694
84 0.0678039714694023
85 0.0665034924944242
86 0.0666604265570641
87 0.0662714739640554
88 0.0651305168867111
89 0.0641998772819837
90 0.063634214301904
91 0.0631707347929478
92 0.0631368309259415
93 0.0629062093794346
94 0.0627623498439789
95 0.0620053360859553
96 0.0614702304204305
97 0.0609723553061485
98 0.0608260110020638
99 0.0608016327023506
100 0.0605254073937734
101 0.0602416433393955
102 0.0606317246953646
103 0.060654749472936
104 0.0612051660815875
105 0.0606625539561113
106 0.0610441515843074
107 0.0619301795959473
108 0.0647881502906481
109 0.0678930406769117
110 0.0655562505125999
111 0.0703750004371007
112 0.0720720117290815
113 0.0641831668714682
114 0.0653830319643021
115 0.0633993161221345
116 0.0642320985595385
117 0.0615642964839935
118 0.0629713299373786
119 0.0607877473036448
120 0.0618634186685085
121 0.0602625645697117
122 0.0607532573242982
123 0.0607570496698221
124 0.0609732853869597
125 0.0600946818788846
126 0.0599343329668045
127 0.0627696936329206
128 0.0611138865351677
129 0.0592928975820541
130 0.0611208602786064
131 0.0597761968771617
132 0.0577169160048167
133 0.0580874942243099
134 0.0579213214417299
135 0.0564201461772124
136 0.0563899949193001
137 0.0561977500716845
138 0.0561617749432723
139 0.0553983288506667
140 0.055460096647342
141 0.0557470694184303
142 0.0561121205488841
143 0.0552570298314095
144 0.0553727621833483
145 0.056286262969176
146 0.0579047923286756
147 0.0588295919199785
148 0.0579848202566306
149 0.0606455604235331
150 0.0695763329664866
151 0.0661254425843557
152 0.0668824041883151
153 0.0676785732309024
154 0.0653768392900626
155 0.0638323649764061
156 0.0687449723482132
157 0.0687509675820668
158 0.0684719880421956
159 0.063421127696832
160 0.0661065926154455
161 0.0644879539807637
162 0.0621452853083611
163 0.0626338049769402
164 0.0607824151714643
165 0.0610899155338605
166 0.0615598621467749
167 0.0590875893831253
168 0.060576772938172
169 0.0597740411758423
170 0.0587325717012087
171 0.058871985723575
172 0.0573968142271042
173 0.056806160757939
174 0.0570850372314453
175 0.0566115503509839
176 0.0568023311595122
177 0.0562547718485196
178 0.0558554505308469
179 0.0550839615364869
180 0.0558901044229666
181 0.0556628468135993
182 0.0551213659346104
183 0.0557507649064064
184 0.0549059274295966
185 0.0550216746826967
186 0.0552835936347644
187 0.0545558010538419
188 0.0554530409475168
189 0.0554324376086394
190 0.0576814226806164
191 0.0580439753830433
192 0.0591170092423757
193 0.0633316300809383
194 0.0681108782688777
195 0.0640773586928844
196 0.0644354149699211
197 0.0683120638132095
198 0.0598220191895962
199 0.0628944933414459
200 0.0587081760168076
201 0.0589153207838535
202 0.0572132008771102
203 0.0581408304472764
204 0.0565352352956931
205 0.0566428204377492
206 0.0561485973497232
207 0.0550140254199505
208 0.0550231138865153
209 0.0538581771155198
210 0.0530997030436993
211 0.0526614437500636
212 0.0528087864319483
213 0.0526556347807248
214 0.0526947838564714
215 0.0519576147198677
216 0.0517807801564535
217 0.05154849588871
218 0.0520504899322987
219 0.0520281332234542
220 0.0511931329965591
221 0.0517594094077746
222 0.0515469647943974
223 0.0516255423426628
224 0.0517165996134281
225 0.0515832863748074
226 0.0529432507852713
227 0.0540523193776608
228 0.0522752205530802
229 0.0537247049311797
230 0.0580093376338482
231 0.059437188009421
232 0.0570789116124312
233 0.0615262351930141
234 0.0659985865155856
235 0.0611791436870893
236 0.0627467359105746
237 0.0599101297557354
238 0.0602100516359011
239 0.0569256184001764
240 0.0556381319959958
241 0.0550258172055086
242 0.0544368761281172
243 0.0537703335285187
244 0.0534954170385996
245 0.0515825090308984
246 0.051123800377051
247 0.0507418115933736
248 0.0501389342049758
249 0.0495187317331632
250 0.0497470038632552
251 0.0496277809143066
252 0.0489772769312064
253 0.0490559749305248
254 0.0494821506241957
255 0.0489212349057198
256 0.049175971498092
257 0.0501710052291552
258 0.0504184849560261
259 0.0492334725956122
260 0.0510512193044027
261 0.0532358375688394
262 0.0528399149576823
263 0.053986797730128
264 0.0552968333164851
265 0.0575476040442785
266 0.0581937618553638
267 0.0586370204885801
268 0.0566511551539103
269 0.0595089954634508
270 0.0648013936976592
271 0.0629672023157279
272 0.0641253901024659
273 0.0677610114216805
274 0.0609351595242818
275 0.0589346028864384
276 0.0583361610770226
277 0.0560162651042143
278 0.054116861273845
279 0.05387827505668
280 0.0538229222098986
281 0.052976556122303
282 0.0530565256873767
283 0.0512325142820676
284 0.0516457309325536
285 0.0513157000144323
286 0.0527754239737988
287 0.052096039056778
288 0.0528544175128142
289 0.0518109326561292
290 0.0507560955981414
291 0.0513317137956619
292 0.0512921313444773
293 0.0506670897205671
294 0.0508319449921449
295 0.0512657960255941
296 0.0509875006973743
297 0.05095391223828
298 0.0518038297692935
299 0.0524818537135919
300 0.0510387743512789
301 0.0510566296676795
302 0.0540848597884178
303 0.0530700658758481
304 0.0522271643082301
305 0.0537877765794595
306 0.0540754186610381
307 0.051738532880942
308 0.0533008500933647
309 0.0524644330143929
310 0.0505789034068584
311 0.05101136987408
312 0.0496704044441382
313 0.0502292228241762
314 0.0505168115099271
315 0.0508791630466779
316 0.0508873611688614
317 0.0496516227722168
318 0.0512832117577394
319 0.0503028345604738
320 0.0497652528186639
321 0.0503225003679593
322 0.0495085902512074
323 0.0490750869115194
324 0.0492296094695727
325 0.0487305236359437
326 0.049229234457016
327 0.0499007329344749
328 0.0494422974685828
329 0.0497028790414333
330 0.0513273949424426
331 0.0538871164123217
332 0.0521310158073902
333 0.0562963436047236
334 0.0595549481610457
335 0.0553052065273126
336 0.0578654184937477
337 0.0531152275701364
338 0.0526008320351442
339 0.0504890568554401
340 0.0496457802752654
341 0.0516744256019592
342 0.0507515755792459
343 0.0514610335230827
344 0.0520671221117179
345 0.0510819479823112
346 0.0509101301431656
347 0.0512394693990548
348 0.0493956270317237
349 0.0487753550211589
350 0.0486675053834915
351 0.0478844133516153
352 0.0482919650773207
353 0.0487742883463701
354 0.0473093055188656
355 0.0485196846226851
356 0.0476932898163795
357 0.0488774068653584
358 0.0503565706312656
359 0.048541035503149
360 0.0510758397479852
361 0.0504450524846713
362 0.0489539280533791
363 0.0522223189473152
364 0.0493271003166834
365 0.0506863221526146
366 0.0509915910661221
367 0.0569638609886169
368 0.0540854098896186
369 0.0535572978357474
370 0.0570029815038045
371 0.0529476143419743
372 0.0571243564287821
373 0.0531504228711128
374 0.0523021693030993
375 0.050912960122029
376 0.0523579070965449
377 0.0507050131758054
378 0.0519246459007263
379 0.0509741492569447
380 0.0514701940119267
381 0.0501243496934573
382 0.0484035387635231
383 0.0496827401220798
384 0.0487307744721572
385 0.0479863546788692
386 0.0482852285106977
387 0.0475900173187256
388 0.0464964471757412
389 0.0477389208972454
390 0.0500350097815196
391 0.0482696257531643
392 0.0486286009351412
393 0.0516555433471998
394 0.0481641193230947
395 0.0481104639669259
396 0.0516476469735304
397 0.049677183230718
398 0.04906099041303
399 0.0510563837985198
400 0.0495183827976386
401 0.0492580272257328
402 0.0512950730820497
403 0.0492939775188764
404 0.0499001915256182
405 0.0506169771154722
406 0.0489594501753648
407 0.0500682952503363
408 0.0489268054564794
409 0.0489730201661587
410 0.0492597098151843
411 0.0470037783185641
412 0.0485830406347911
413 0.0488720635573069
414 0.0461072189112504
415 0.0483664696415265
416 0.04732346534729
417 0.0468028833468755
418 0.0468219767014186
419 0.0461079714198907
420 0.0456070490181446
421 0.0453310136993726
422 0.0457817229131858
423 0.0451044080158075
424 0.0454163812100887
425 0.0456193760037422
426 0.0464807425936063
427 0.046430795143048
428 0.0467301321526368
429 0.0534455912808577
430 0.0547348211208979
431 0.0516575624545415
432 0.0588338325421015
433 0.0520377246042093
434 0.0562319159507751
435 0.0562776376803716
436 0.0561491387585799
437 0.0597884853680929
438 0.0647632541755835
439 0.0622166829804579
440 0.0672003937264283
441 0.0599070452153683
442 0.0581394011775653
443 0.0559891487161318
444 0.0537050428489844
445 0.0541561270753543
446 0.0505906989177068
447 0.0516203008592129
448 0.0514042650659879
449 0.0500679798424244
450 0.0485159978270531
451 0.0486268711586793
452 0.0491519048810005
453 0.0471281905968984
454 0.0469393817087015
455 0.0477096637090047
456 0.0474595986306667
457 0.0469064079225063
458 0.0462706858913104
459 0.0460291827718417
460 0.0448949038982391
461 0.0447323682407538
462 0.0440578845640024
463 0.0441619865596294
464 0.043424674620231
465 0.0435769396523635
466 0.043658934533596
467 0.0434740756948789
468 0.0433480963110924
469 0.0434296702345212
470 0.0427388176321983
471 0.0428517137964567
472 0.0425076459844907
473 0.0424522211154302
474 0.0421404267350833
475 0.0423179628948371
476 0.0429260159532229
477 0.0427402208248774
478 0.0428906343877316
479 0.0433894731104374
480 0.0443813515206178
481 0.0465917078157266
482 0.0476276824871699
483 0.0503438872595628
484 0.0551205811401208
485 0.0651488626996676
486 0.0649874806404114
487 0.0639152253667514
488 0.0658257889250914
489 0.0565480142831802
490 0.0647906089822451
491 0.0572238514820735
492 0.0563424502809842
493 0.0570657315353552
494 0.0526687192420165
495 0.0527741983532906
496 0.0514913387596607
497 0.0488186776638031
498 0.0480353621145089
499 0.0482752832273642
500 0.0466392214099566
501 0.0460193169613679
502 0.0473503607014815
503 0.0453936842580636
504 0.046614001194636
505 0.0466130003333092
506 0.0463853615025679
507 0.0469680577516556
508 0.0455940750737985
509 0.0467894288400809
510 0.0457014751931032
511 0.0456140401462714
512 0.0457827374339104
513 0.0454801581799984
514 0.0465540649990241
515 0.0457208553949992
516 0.046594695498546
517 0.0455126749972502
518 0.0467503592371941
519 0.0462327686448892
520 0.0460335239768028
521 0.0475789085030556
522 0.046720627695322
523 0.0489724365373453
524 0.0476048700511456
525 0.0484198642273744
526 0.0485940811534723
527 0.0484434502820174
528 0.051658172160387
529 0.0498125180602074
530 0.0510817244648933
531 0.0521118442217509
532 0.051642627765735
533 0.0528511454661687
534 0.0517008379101753
535 0.055308073759079
536 0.0516382729013761
537 0.0526566170156002
538 0.0559655539691448
539 0.0523257305224737
540 0.055146483083566
541 0.0538080036640167
542 0.050827136884133
543 0.0502539463341236
544 0.0511524652441343
545 0.0500648928185304
546 0.0521902019778887
547 0.0526450412968794
548 0.0501363761723042
549 0.0484139273564021
550 0.0482196944455306
551 0.0479837519427141
552 0.0472413289050261
553 0.050148411343495
554 0.0518130858739217
555 0.0517010912299156
556 0.0545744014283021
557 0.0551891811192036
558 0.0533121377229691
559 0.0505576692521572
560 0.0489001087844372
561 0.0492518519361814
562 0.049539594600598
563 0.0505183897912502
564 0.0530722501377265
565 0.0541476346552372
566 0.053617258866628
567 0.0544567592442036
568 0.0534251642723878
569 0.0498503521084785
570 0.0500006899237633
571 0.0497409142553806
572 0.05126928165555
573 0.0500713996589184
574 0.0488492250442505
575 0.046597383916378
576 0.0451213965813319
577 0.045332457870245
578 0.0454146737853686
579 0.0446274789671103
580 0.0451622431476911
581 0.0450726573665937
582 0.0449266793827216
583 0.0448331932226817
584 0.0443612759311994
585 0.0444574443002542
586 0.0457198744018873
587 0.0451738809545835
588 0.0467166056235631
589 0.0502746974428495
590 0.0481100852290789
591 0.0492112350960573
592 0.0516546269257863
593 0.0467032492160797
594 0.0487850358088811
595 0.0470775129894416
596 0.0462759571770827
597 0.0463358958562215
598 0.0462796141703924
599 0.0455703896780809
600 0.0462490680317084
601 0.046624888976415
602 0.0470343592266242
603 0.0470488394300143
604 0.0468161093691985
605 0.0481341357032458
606 0.0466405140856902
607 0.046287061025699
608 0.04506499816974
609 0.0454457836846511
610 0.0450694113969803
611 0.0450385225315889
612 0.0455851716299852
613 0.0450083091855049
614 0.0446047273774942
615 0.0442315277953943
616 0.0432600937783718
617 0.0427513631681601
618 0.0421275633076827
619 0.0428332574665546
620 0.0420953209201495
621 0.0423809587955475
622 0.0421576015651226
623 0.0420135433475176
624 0.0413741531471411
625 0.0427185681958993
626 0.0426579167445501
627 0.0423898324370384
628 0.0437327511608601
629 0.0465403186778227
630 0.046747633566459
631 0.0462722691396872
632 0.0512179397046566
633 0.0488861898581187
634 0.0471848584711552
635 0.05296366289258
636 0.0468458893398444
637 0.0502383075654507
638 0.0493098013103008
639 0.0483127261201541
640 0.0502931115527948
641 0.0495594268043836
642 0.0492765928308169
643 0.0485568729539712
644 0.046821691095829
645 0.0485657788813114
646 0.0465401224792004
647 0.0458760571976503
648 0.0453474496801694
649 0.044616070886453
650 0.0455923366049925
651 0.044549665103356
652 0.0460079225401084
653 0.0439734123647213
654 0.0456676793595155
655 0.0440506637096405
656 0.0465580609937509
657 0.0448667928576469
658 0.0452822633087635
659 0.0450987269481023
660 0.0457065068185329
661 0.0448723373313745
662 0.0454728665451209
663 0.045757669955492
664 0.0454636203746001
665 0.0444376443823179
666 0.0439016843835513
667 0.044342594842116
668 0.0445737602810065
669 0.0437855906784534
670 0.0463678787151972
671 0.0457057803869247
672 0.0460336940983931
673 0.0473431671659152
674 0.046781616906325
675 0.0464334438244502
676 0.0468777095278104
677 0.0476448660095533
678 0.0476475767791271
679 0.0511490864058336
680 0.0485088403026263
681 0.0512141498426596
682 0.0505587061246236
683 0.0494019128382206
684 0.049194261431694
685 0.0495906608800093
686 0.0482975132763386
687 0.0483272324005763
688 0.0466771759092808
689 0.0487463511526585
690 0.0470988489687443
691 0.0477995499968529
692 0.0483007145424684
693 0.0478228492041429
694 0.0498276961346467
695 0.0486003247400125
696 0.0497339330613613
697 0.0500989270706972
698 0.0521367577215036
699 0.0535022877156734
700 0.0543240333596865
701 0.0560284852981567
702 0.0533285414179166
703 0.0518398694694042
704 0.0545192807912827
705 0.053967093427976
706 0.0550403694311778
707 0.0508688315749168
708 0.0498952865600586
709 0.0462375742693742
710 0.0459293921788534
711 0.0460613258183002
712 0.0453586714963118
713 0.0460885527233283
714 0.0454739555716515
715 0.0459158644080162
716 0.0459566650291284
717 0.0437839888036251
718 0.0438332557678223
719 0.0433291892210642
720 0.0438684746623039
721 0.0435569124917189
722 0.0438555267949899
723 0.0436500559250514
724 0.0434490516781807
725 0.0431306436657906
726 0.0426030258337657
727 0.0428049452602863
728 0.0418325861295064
729 0.0420759531358878
730 0.0424340727428595
731 0.0412043357888858
732 0.0423302451769511
733 0.0420635665456454
734 0.0421830030779044
735 0.0422298386693001
736 0.0433522164821625
737 0.042424600571394
738 0.0428282891710599
739 0.0441181734204292
740 0.0430299863219261
741 0.0448227922121684
742 0.0459833443164825
743 0.04584543282787
744 0.0456730810304483
745 0.049214789023002
746 0.0471099962790807
747 0.0494071207940578
748 0.0496829027930895
749 0.0478566760818164
750 0.0476807281374931
751 0.0490038841962814
752 0.050569724291563
753 0.050611204157273
754 0.0518433749675751
755 0.0506839267909527
756 0.0529029866059621
757 0.0482282216350238
758 0.0488565353055795
759 0.0471368047098319
760 0.0462453824778398
761 0.0450748801231384
762 0.0448992773890495
763 0.0440770387649536
764 0.0427121308942636
765 0.0426196530461311
766 0.0432419652740161
767 0.0426696203649044
768 0.0425034860769908
769 0.0417759132881959
770 0.0413762591779232
771 0.0410709194839001
772 0.0410937555134296
773 0.0420518703758717
774 0.0409434884786606
775 0.0411417682965597
776 0.0411627826591333
777 0.0421452547113101
778 0.0415018896261851
779 0.0426744520664215
780 0.0424284115433693
781 0.0423988861342271
782 0.0419186403354009
783 0.042917400598526
784 0.0437298814455668
785 0.0440572425723076
786 0.0451023752490679
787 0.0443215717871984
788 0.0450654874245326
789 0.0438137066860994
790 0.0450706606109937
791 0.0444361517826716
792 0.0452757266660531
793 0.0449141773084799
794 0.046960923820734
795 0.0457297066847483
796 0.0479265774289767
797 0.0482576203842958
798 0.0466256774961948
799 0.0494121188918749
800 0.0463122762739658
801 0.0499892433484395
802 0.0465961756805579
803 0.0496657515565554
804 0.0461754513283571
805 0.0476552943388621
806 0.0474320041636626
807 0.0447030439972878
808 0.0485147746900717
809 0.0457173138856888
810 0.0463452314337095
811 0.0477439475556215
812 0.0451072504123052
813 0.0487065687775612
814 0.0454481095075607
815 0.0437758763631185
816 0.0476260520517826
817 0.0450446593264739
818 0.0442829951643944
819 0.0464431221286456
820 0.0429235150416692
821 0.0435809679329395
822 0.0444812501470248
823 0.0432352721691132
824 0.0447917257746061
825 0.0472878478467464
826 0.0455429057280223
827 0.0481318173309167
828 0.049346453199784
829 0.0476091355085373
830 0.0490753253300985
831 0.0495622915526231
832 0.0452522026995818
833 0.0462801344692707
834 0.0478526850541433
835 0.0467425572375456
836 0.0464704794188341
837 0.0469142595926921
838 0.0460636305312316
839 0.0453627842168013
840 0.0443561151623726
841 0.0438343522449334
842 0.0420460800329844
843 0.0418229947487513
844 0.0425297766923904
845 0.0422944587965806
846 0.0432724493245284
847 0.042504571378231
848 0.0438320289055506
849 0.0432822604974111
850 0.0439538458983103
851 0.0449155767758687
852 0.0440453576544921
853 0.0447196351985137
854 0.0450545872251193
855 0.0439941262205442
856 0.0445742805798848
857 0.0442226740221182
858 0.043800978610913
859 0.0455501824617386
860 0.0446476029853026
861 0.0459111568828424
862 0.046909953157107
863 0.0471025755008062
864 0.0498237883051236
865 0.0469948177536329
866 0.0481943835814794
867 0.0466855888565381
868 0.0455784474809965
869 0.0440551117062569
870 0.0442932558556398
871 0.0453479712208112
872 0.0449196845293045
873 0.0446534814933936
874 0.04360090320309
875 0.0426757708191872
876 0.0421941814323266
877 0.042272159208854
878 0.0413270468513171
879 0.0428601813813051
880 0.0420287462572257
881 0.041190005838871
882 0.041907723993063
883 0.0408529229462147
884 0.0404528851310412
885 0.0405318513512611
886 0.0399711094796658
887 0.0398237444460392
888 0.0407609902322292
889 0.0402041884760062
890 0.039813739558061
891 0.0414144955575466
892 0.039918045202891
893 0.040539229909579
894 0.0425010286271572
895 0.0414249238868554
896 0.0418129911025365
897 0.0431871227920055
898 0.0424858431021372
899 0.0426595285534859
900 0.0459639839828014
901 0.0443123529354731
902 0.0465113433698813
903 0.0481052050987879
904 0.0450559469560782
905 0.0502237925926844
906 0.0471318985025088
907 0.0488998703658581
908 0.0481121378640334
909 0.0492675118148327
910 0.0501611654957136
911 0.0492387413978577
912 0.0493333699802558
913 0.0506702202061812
914 0.0499259829521179
915 0.0521582153936227
916 0.0491287087400754
917 0.0487726231416067
918 0.0460629810889562
919 0.0449783677856127
920 0.045346866051356
921 0.0439816998938719
922 0.0434432079394658
923 0.0426352210342884
924 0.0410848806301753
925 0.0420263906319936
926 0.0411217312018077
927 0.0418811527391275
928 0.0424188288549582
929 0.0422486712535222
930 0.0421277433633804
931 0.0411591194570065
932 0.0427115696171919
933 0.0421720830102762
934 0.0416775892178218
935 0.0426239830752214
936 0.0421755649149418
937 0.0419893289605776
938 0.0416982645789782
939 0.0420432686805725
940 0.0439119748771191
941 0.0425804778933525
942 0.0442542371650537
943 0.044154287626346
944 0.0427691278358301
945 0.0451359711587429
946 0.0442383550107479
947 0.0449758147199949
948 0.0431308994690577
949 0.0428472384810448
950 0.0442198614279429
951 0.0423174512883027
952 0.0460247111817201
953 0.0455642528831959
954 0.0459915859003862
955 0.0470782270034154
956 0.0449526992936929
957 0.0470009793837865
958 0.0429850642879804
959 0.043473510692517
960 0.0448259587089221
961 0.0433244196077188
962 0.0456634648144245
963 0.0427793823182583
964 0.0431185849010944
965 0.0424847876032194
966 0.0411358972390493
967 0.0416848833362261
968 0.0417979955673218
969 0.041177898645401
970 0.0411284665266673
971 0.0399449480076631
972 0.0402442701160908
973 0.0395866545538108
974 0.040201428035895
975 0.0405988097190857
976 0.0416817354659239
977 0.0412895232439041
978 0.0413435610632102
979 0.0411052753527959
980 0.0407927359143893
981 0.0411877358953158
982 0.0419386165837447
983 0.0425333939492702
984 0.0433118517200152
985 0.0446919836103916
986 0.0437051380674044
987 0.0439696038762728
988 0.0436025224626064
989 0.0428967910508315
990 0.0454525959988435
991 0.0441537176569303
992 0.0457565685113271
993 0.046885156383117
994 0.0473986826837063
995 0.0473555425802867
996 0.0484852654238542
997 0.049326969931523
998 0.0477077302833398
999 0.0482710041105747
1000 0.0474271016816298
1001 0.0445289984345436
1002 0.0474479235708714
1003 0.0457394098242124
1004 0.0455646887421608
1005 0.0453008090456327
1006 0.0443562157452106
1007 0.0427716262638569
1008 0.0429572897652785
1009 0.0418469694753488
1010 0.0417055276532968
1011 0.0408577546477318
1012 0.041174237926801
1013 0.0405720459918181
1014 0.0409617051482201
1015 0.0404460057616234
1016 0.0405867459873358
1017 0.0402693611880144
1018 0.0414587234457334
1019 0.0402479792634646
1020 0.0407999344170094
1021 0.0398717063168685
1022 0.0412725321948528
1023 0.0399875541528066
1024 0.0406738743185997
1025 0.0408867597579956
1026 0.0408203353484472
1027 0.0420126294096311
1028 0.0417742170393467
1029 0.0426439742247264
1030 0.041875367363294
1031 0.0419856905937195
1032 0.0423218496143818
1033 0.0420360192656517
1034 0.0427555255591869
1035 0.043639350682497
1036 0.0445253973205884
1037 0.0443688171605269
1038 0.0439796857535839
1039 0.0434746332466602
1040 0.0431274548172951
1041 0.0422654437522093
1042 0.0422864034771919
1043 0.0433800878624121
1044 0.0427428061763446
1045 0.0432596852382024
1046 0.0439129248261452
1047 0.0426259438196818
1048 0.0427548363804817
1049 0.042042816678683
1050 0.0417120059331258
1051 0.041778888553381
1052 0.0435032894213994
1053 0.0425369900961717
1054 0.0438518635928631
1055 0.0457462506989638
1056 0.0436745931704839
1057 0.04842742284139
1058 0.0441875606775284
1059 0.0470244698226452
1060 0.046057050426801
1061 0.0455674057205518
1062 0.0475648070375125
1063 0.0443937219679356
1064 0.046448548634847
1065 0.0482398420572281
1066 0.0444266920288404
1067 0.0479023791849613
1068 0.0474809072911739
1069 0.0444937311112881
1070 0.044772798816363
1071 0.0434165187180042
1072 0.0417111255228519
1073 0.0431165980796019
1074 0.0426949970424175
1075 0.0408878537515799
1076 0.044036856542031
1077 0.0441233466068904
1078 0.041812806079785
1079 0.0411653332412243
1080 0.0417488192518552
1081 0.0416658533116182
1082 0.0421267151832581
1083 0.0425996892154217
1084 0.0413497338692347
1085 0.0428246545294921
1086 0.0433445324500402
1087 0.0413093542059263
1088 0.0423004229863485
1089 0.0436539227763812
1090 0.0409423088033994
1091 0.0426224991679192
1092 0.0422226190567017
1093 0.0406138027707736
1094 0.0421658443907897
1095 0.0417545313636462
1096 0.0394005912045638
1097 0.0423617338140806
1098 0.0431609402100245
1099 0.0410655091206233
1100 0.0426699779927731
1101 0.0440839839478334
1102 0.041682085643212
1103 0.0426646197835604
1104 0.0438753329217434
1105 0.0420697716375192
1106 0.0437149194379648
1107 0.0464998135964076
1108 0.0445548730591933
1109 0.0444446168839931
1110 0.046158788104852
1111 0.0450269331534704
1112 0.042548185835282
1113 0.0448195300996304
1114 0.0462329412500064
1115 0.043299230436484
1116 0.0455101020634174
1117 0.0467255264520645
1118 0.0454211235046387
1119 0.0458177253603935
1120 0.0440522966285547
1121 0.0473932015399138
1122 0.0472541625301043
1123 0.0458539637426535
1124 0.0462353900074959
1125 0.0472798819343249
1126 0.046935665110747
1127 0.0462373880048593
1128 0.0459690776964029
1129 0.0500096157193184
1130 0.0531337869664033
1131 0.0574555819233259
1132 0.0566927629212538
1133 0.0532750524580479
1134 0.0533818105856578
1135 0.0540373437106609
1136 0.0525948305924734
1137 0.0513722933828831
1138 0.0498697919150194
1139 0.048011302947998
1140 0.0482143921156724
1141 0.0470106986661752
1142 0.0476442240178585
1143 0.0452208817005157
1144 0.0446101625760396
1145 0.0432887226343155
1146 0.0426593013107777
1147 0.0415499098598957
1148 0.0415404997766018
1149 0.0405751690268517
1150 0.040108231206735
1151 0.0397762830058734
1152 0.0398468139270941
1153 0.0394524211684863
1154 0.0392953244348367
1155 0.038778784374396
1156 0.0387747151156267
1157 0.0387173866232236
1158 0.0388669135669867
1159 0.0387170786658923
1160 0.0393647452195485
1161 0.0393888379136721
1162 0.0396496616303921
1163 0.0395261943340302
1164 0.0399133861064911
1165 0.0398590142528216
1166 0.0405763685703278
1167 0.0408041055003802
1168 0.0416962578892708
1169 0.0414698968331019
1170 0.0423624031245708
1171 0.0422321570416292
1172 0.0425202722350756
1173 0.042341373860836
1174 0.04279750213027
1175 0.0425267790754636
1176 0.0429406302670638
1177 0.0427375646928946
1178 0.0420964869360129
1179 0.0417908368011316
1180 0.0406243316829205
1181 0.0401724229256312
1182 0.0393199411531289
1183 0.0389874515434106
1184 0.0383906041582425
1185 0.0385670575002829
1186 0.0379780307412148
1187 0.0381612492104371
1188 0.0378664098680019
1189 0.0376314073801041
1190 0.0379264578223228
1191 0.0379165187478065
1192 0.0374065848688285
1193 0.038714911788702
1194 0.038221962749958
1195 0.0377291006346544
1196 0.0389043254156907
1197 0.0383149472375711
1198 0.0382188521325588
1199 0.0396216983596484
1200 0.0383941692610582
1201 0.0396314139167468
1202 0.0395328464607398
1203 0.0400506916145484
1204 0.0410224062701066
1205 0.0400267193714778
1206 0.0431278658409913
1207 0.0411995661755403
1208 0.0423178039491177
1209 0.0409150359531244
1210 0.0414797179400921
1211 0.0415578844646613
1212 0.0410787040988604
1213 0.0438185321787993
1214 0.0414921008050442
1215 0.0440363486607869
1216 0.0418373557428519
1217 0.0445063089330991
1218 0.0444752698143323
1219 0.0465469397604465
1220 0.0458380418519179
1221 0.0466463727255662
1222 0.0475883384545644
1223 0.0534070730209351
1224 0.0547927344838778
1225 0.0522510533531507
1226 0.0491482069094976
1227 0.0459015543262164
1228 0.0443171970546246
1229 0.0435769744217396
1230 0.0435186363756657
1231 0.0430750114222368
1232 0.0426370278000832
1233 0.0416443385183811
1234 0.0417081539829572
1235 0.0395233556628227
1236 0.0393265523016453
1237 0.0388005922238032
1238 0.0385158210992813
1239 0.0386969124277433
1240 0.0382696328063806
1241 0.0373738370835781
1242 0.038101390004158
1243 0.0372815256317457
1244 0.0369692767659823
1245 0.0373299059768518
1246 0.0367985777556896
1247 0.0370495580136776
1248 0.0374401894708475
1249 0.0370552539825439
1250 0.0371865717073282
1251 0.0371983088552952
1252 0.0372886521120866
1253 0.0376270711421967
1254 0.0374850196142991
1255 0.0375195145606995
1256 0.0382332801818848
1257 0.038691279788812
1258 0.0387789408365885
1259 0.0399451081951459
1260 0.0397258053223292
1261 0.0402211621403694
1262 0.0402135836581389
1263 0.0407652693490187
1264 0.0409109008808931
1265 0.041998352855444
1266 0.0423193077246348
1267 0.0416424075762431
1268 0.0434851261476676
1269 0.0443439843753974
1270 0.0455591479937235
1271 0.0479920754830042
1272 0.0468529586990674
1273 0.044356070458889
1274 0.0419487282633781
1275 0.0394277945160866
1276 0.0393040180206299
1277 0.0394636889298757
1278 0.0388774064679941
1279 0.0390830958882968
1280 0.0390678706268469
1281 0.0387032901247342
1282 0.0403686886032422
1283 0.0397195567687353
1284 0.0389235975841681
1285 0.0391719453036785
1286 0.0380655551950137
1287 0.037134392807881
1288 0.0369517679015795
1289 0.0374767469863097
1290 0.0368739478290081
1291 0.037722942729791
1292 0.0378602594137192
1293 0.0367169827222824
1294 0.0386481992900372
1295 0.038756001740694
1296 0.0380009114742279
1297 0.0400863314668337
1298 0.0383611892660459
1299 0.0385026236375173
1300 0.0410756766796112
1301 0.0392614727218946
1302 0.0422925315797329
1303 0.040435458223025
1304 0.0405211299657822
1305 0.0420856438577175
1306 0.0404455289244652
1307 0.0430668083329995
1308 0.0419209823012352
1309 0.0432455589373906
1310 0.0460337201754252
1311 0.0472460227708022
1312 0.0472189995149771
1313 0.0476939516762892
1314 0.0456447328130404
1315 0.0463730456928412
1316 0.0447057820856571
1317 0.0463615544140339
1318 0.0470374288658301
1319 0.0473059453070164
1320 0.0464878814915816
1321 0.0445568983753522
1322 0.0418917474647363
1323 0.040861318508784
1324 0.0405116615196069
1325 0.0402060759564241
1326 0.0393113953371843
1327 0.0412057874103387
1328 0.0405657204488913
1329 0.0416837086280187
1330 0.0408568394680818
1331 0.041094932705164
1332 0.0400766643385092
1333 0.0388078925510248
1334 0.0404691162208716
1335 0.0390633568167686
1336 0.0406744678815206
1337 0.0401708744466305
1338 0.0417207839588324
1339 0.0417530785004298
1340 0.0423202700912952
1341 0.0414112371702989
1342 0.0408163989583651
1343 0.0402292981743813
1344 0.0403089622656504
1345 0.0394999732573827
1346 0.040523869295915
1347 0.0415349813799063
1348 0.0419535674154758
1349 0.0404591262340546
1350 0.0406829454004765
1351 0.04152953500549
1352 0.0401210784912109
1353 0.0426266913612684
1354 0.0418708051244418
1355 0.0419967944423358
1356 0.0407616930703322
1357 0.0408169204990069
1358 0.0404073260724545
1359 0.0403690338134766
1360 0.0397141439219316
1361 0.040117704619964
1362 0.0411182095607122
1363 0.0395702595512072
1364 0.0396709938844045
1365 0.0389901163677374
1366 0.0402379731337229
1367 0.0401623410483201
1368 0.039277778317531
1369 0.0397224376598994
1370 0.0390526180466016
1371 0.0387879461050034
1372 0.0387496339778105
1373 0.0381539451579253
1374 0.0372343957424164
1375 0.0372713580727577
1376 0.0376250483095646
1377 0.0383643743892511
1378 0.0387676867345969
1379 0.0388049396375815
1380 0.0388050340116024
1381 0.0385309817890326
1382 0.0377284561594327
1383 0.0399424334367116
1384 0.0413972822328409
1385 0.0413769260048866
1386 0.0434660042325656
1387 0.0431211603184541
1388 0.0424561202526093
1389 0.0414855256676674
1390 0.0428715112308661
1391 0.0449099875986576
1392 0.0463927909731865
1393 0.0471540863315264
1394 0.0473676783343156
1395 0.0465675505499045
1396 0.0446968302130699
1397 0.0451964723567168
1398 0.0442868831257025
1399 0.0448988750576973
1400 0.0429978221654892
1401 0.0409061834216118
1402 0.0413198905686537
1403 0.0409902222454548
1404 0.0390356319646041
1405 0.0390072862307231
1406 0.0388390632967154
1407 0.0380337896446387
1408 0.0382936584452788
1409 0.036805355300506
1410 0.0365671788652738
1411 0.0371967032551765
1412 0.0363565062483152
1413 0.036524744083484
1414 0.038315761834383
1415 0.0364028563102086
1416 0.0374681986868382
1417 0.0373227198918661
1418 0.0370129980146885
1419 0.0392324316004912
1420 0.0374411356945833
1421 0.0391872984667619
1422 0.0411349435647329
1423 0.0390484804908435
1424 0.0411499552428722
1425 0.0390141904354095
1426 0.0388892504076163
1427 0.0402004569768906
1428 0.039307684948047
1429 0.0393976954122384
1430 0.0423292778432369
1431 0.0401505070428054
1432 0.0413724879423777
1433 0.0426456953088442
1434 0.040449246764183
1435 0.0436314195394516
1436 0.0424603472153346
1437 0.0399669508139292
1438 0.0429076527555784
1439 0.0419398347536723
1440 0.0406554614504178
1441 0.0412665059169134
1442 0.0424321964383125
1443 0.0425944825013479
1444 0.0462509008745352
1445 0.0492758626739184
1446 0.0459036082029343
1447 0.0460965198775133
1448 0.0507648835579554
1449 0.0483180532852809
1450 0.0458002872765064
1451 0.0459276052812735
1452 0.046469288567702
1453 0.0460168806215127
1454 0.0456162144740423
1455 0.0437801294028759
1456 0.0437765630582968
1457 0.0469592300554117
1458 0.0497657147546609
1459 0.0477094240486622
1460 0.0470989073316256
1461 0.0471805321673552
1462 0.0477600681285063
1463 0.0456139383216699
1464 0.0439931924144427
1465 0.0435326260824998
1466 0.0422526200612386
1467 0.041802067309618
1468 0.042249895632267
1469 0.0415722876787186
1470 0.0404601780076822
1471 0.0404478994508584
1472 0.0401318954924742
1473 0.0395553236206373
1474 0.0392527754108111
1475 0.0385700749854247
1476 0.0366984208424886
1477 0.0369449419279893
1478 0.0367273564140002
1479 0.0359369752307733
1480 0.0360039944450061
1481 0.0353388749063015
1482 0.0355078925689062
1483 0.0348416306078434
1484 0.0352365200718244
1485 0.035023837039868
1486 0.0344327365358671
1487 0.0350858593980471
1488 0.0346805118024349
1489 0.034783819069465
1490 0.0354199645419916
1491 0.0347223617136478
1492 0.0356088727712631
1493 0.0362314420441786
1494 0.0357357288400332
1495 0.0371064196030299
1496 0.0368044798572858
1497 0.0365235917270184
1498 0.0379613749682903
1499 0.0373832061886787
1500 0.0386106198032697
1501 0.0387853210171064
1502 0.0389516291519006
1503 0.0407412722706795
1504 0.040143396705389
1505 0.0412933876117071
1506 0.0404174911479155
1507 0.0418865196406841
1508 0.0409034403661887
1509 0.0412037807206313
1510 0.0426832176744938
1511 0.0420181266963482
1512 0.0429071734348933
1513 0.0445213330288728
1514 0.0442687273025513
1515 0.0443274093170961
1516 0.0491112880408764
1517 0.0463773409525553
1518 0.0429490047196547
1519 0.0456379589935144
1520 0.0462281505266825
1521 0.0450334474444389
1522 0.0451559349894524
1523 0.0475126753250758
1524 0.0459774533907572
1525 0.0454816408455372
1526 0.0442206611235937
1527 0.0424462432662646
1528 0.0430790049334367
1529 0.0444384900232156
1530 0.0428729057312012
1531 0.041839711368084
1532 0.042181475708882
1533 0.0422830867270629
1534 0.041335745404164
1535 0.0402161243061225
1536 0.0388800799846649
1537 0.0389902840058009
1538 0.0381884649395943
1539 0.0380578227341175
1540 0.0365473876396815
1541 0.0365579947829247
1542 0.0372584102054437
1543 0.0385290222863356
1544 0.0384130366146564
1545 0.0389720387756824
1546 0.0395311663548152
1547 0.0398417947192987
1548 0.0393947325646877
1549 0.0388134345412254
1550 0.038472980260849
1551 0.0395506843924522
1552 0.0406477811435858
1553 0.0416348117093245
1554 0.0425345872839292
1555 0.04294561718901
1556 0.0429208849867185
1557 0.0418718295792739
1558 0.0408755391836166
1559 0.0389073267579079
1560 0.0383788732190927
1561 0.0375771783292294
1562 0.0367874639729659
1563 0.0359520502388477
1564 0.0358066782355309
1565 0.0353043600916862
1566 0.0350906389455001
1567 0.0351192529002825
1568 0.0349996536970139
1569 0.035046444584926
1570 0.0352161514262358
1571 0.0355030770103137
1572 0.0358331886430581
1573 0.0364378566543261
1574 0.0370907088120778
1575 0.0372958593070507
1576 0.0387202662726243
1577 0.0395122244954109
1578 0.0397993450363477
1579 0.0415317292014758
1580 0.0421691909432411
1581 0.0428766806920369
1582 0.044857910523812
1583 0.0449302966396014
1584 0.0444014507035414
1585 0.045728641251723
1586 0.0451756777862708
1587 0.0453522888322671
1588 0.0442171916365623
1589 0.0445695829888185
1590 0.0421846186121305
1591 0.041107897957166
1592 0.0402273287375768
1593 0.0402165042857329
1594 0.0390575056274732
1595 0.0383293939133485
1596 0.0372661724686623
1597 0.0376741699874401
1598 0.0376700696845849
1599 0.0373586329321067
1600 0.0372090588013331
1601 0.0375383620460828
1602 0.0381195383767287
1603 0.0383307884136836
1604 0.0388593350847562
1605 0.0378557480871677
1606 0.0378036759793758
1607 0.0377500069638093
1608 0.0373447090387344
1609 0.0377736960848173
1610 0.0375141414503256
1611 0.0365241045753161
1612 0.0366782049338023
1613 0.0356921988228957
1614 0.0357382992903391
1615 0.0356726013123989
1616 0.0361008035639922
1617 0.035435556123654
1618 0.0360100157558918
1619 0.0354299296935399
1620 0.03606291487813
1621 0.0361368134617805
1622 0.0355591488381227
1623 0.0362797441581885
1624 0.0358802204330762
1625 0.0360695645213127
1626 0.0370699192086856
1627 0.0362603602310022
1628 0.0363175297776858
1629 0.0376833577950796
1630 0.0378372532625993
1631 0.0375311238070329
1632 0.0376826263964176
1633 0.0384029534955819
1634 0.0397726222872734
1635 0.0408797487616539
1636 0.0430929486950239
1637 0.043265006194512
1638 0.0418611938754717
1639 0.041297851751248
1640 0.039224615941445
1641 0.037477554132541
1642 0.038531723121802
1643 0.0383938116331895
1644 0.0377251195410887
1645 0.0383101428548495
1646 0.0374718966583411
1647 0.038042905429999
1648 0.0366644921402136
1649 0.0381351448595524
1650 0.0380667584637801
1651 0.0386570766568184
1652 0.0395935513079166
1653 0.0400634184479713
1654 0.0393600128591061
1655 0.0390813338259856
1656 0.0388637197514375
1657 0.0388495177030563
1658 0.03817010546724
1659 0.0395624948044618
1660 0.0386672653257847
1661 0.0401472536226114
1662 0.0390396478275458
1663 0.0379261250297228
1664 0.0370568136374156
1665 0.0365863492091497
1666 0.0364102174838384
1667 0.0369891536732515
1668 0.0389975259701411
1669 0.0386056390901407
1670 0.040249940007925
1671 0.0410719575981299
1672 0.0397124650577704
1673 0.0405457603434722
1674 0.0402820048232873
1675 0.041599053889513
1676 0.0407513678073883
1677 0.0403893378873666
1678 0.0404247802992662
1679 0.0415222669641177
1680 0.0402308491369089
1681 0.0401343752940496
1682 0.0413517902294795
1683 0.0407329301039378
1684 0.0425573078294595
1685 0.0391024400790532
1686 0.0419159916539987
1687 0.0390303333600362
1688 0.0385249257087708
1689 0.0387855718533198
1690 0.0387752577662468
1691 0.0376803502440453
1692 0.0383023930092653
1693 0.0395517597595851
1694 0.0389064190288385
1695 0.0421578971048196
1696 0.0395930781960487
1697 0.0385502601663272
1698 0.0393455550074577
1699 0.0383453071117401
1700 0.0361996876696746
1701 0.0385374464094639
1702 0.039328146725893
1703 0.0386170372366905
1704 0.0401805030802886
1705 0.0396256682773431
1706 0.0375731227298578
1707 0.0391197639207045
1708 0.0403749706844489
1709 0.0378359829386075
1710 0.0384666981796424
1711 0.0410026870667934
1712 0.0404815499981244
1713 0.0380952022969723
1714 0.0402905071775118
1715 0.0416119309763114
1716 0.039680855969588
1717 0.0402411880592505
1718 0.042639534920454
1719 0.043554666141669
1720 0.0422750649352868
1721 0.0423400290310383
1722 0.0432746397952239
1723 0.0461578145623207
1724 0.0470945946872234
1725 0.0451344226797422
1726 0.0434867578248183
1727 0.0440077371895313
1728 0.0440288955966632
1729 0.0439483188092709
1730 0.0426605368653933
1731 0.0414928036431471
1732 0.0405785838762919
1733 0.0383230571945508
1734 0.0371125005185604
1735 0.0359788201749325
1736 0.0360759260753791
1737 0.0363225440184275
1738 0.0362675028542678
1739 0.0358918160200119
1740 0.0350728357831637
1741 0.0345127247273922
1742 0.0341259663303693
1743 0.0341120362281799
1744 0.0340094740192095
1745 0.0342563837766647
1746 0.034088097512722
1747 0.0343131857613722
1748 0.0350300533076127
1749 0.0349247542520364
1750 0.0354117564857006
1751 0.0361182664831479
1752 0.0356299082438151
1753 0.0364755305151145
1754 0.0374260221918424
1755 0.0371563211083412
1756 0.0387339629232883
1757 0.0407093750933806
1758 0.039998434484005
1759 0.0409508359928926
1760 0.0398915956417719
1761 0.0392190714677175
1762 0.0402912398179372
1763 0.0406865018109481
1764 0.042860912779967
1765 0.0427837396661441
1766 0.0429397809008757
1767 0.0426618345081806
1768 0.0402502988775571
1769 0.0407441966235638
1770 0.0403821095824242
1771 0.0404170006513596
1772 0.0397868417203426
1773 0.0391761722664038
1774 0.0376837067306042
1775 0.0378109936912855
1776 0.0368399942914645
1777 0.0368308785061041
1778 0.0358054389556249
1779 0.035940640916427
1780 0.0360530925293764
1781 0.0366276080409686
1782 0.0362070351839066
1783 0.0357633841534456
1784 0.0366200655698776
1785 0.0365451810260614
1786 0.0358917191624641
1787 0.0367634234329065
1788 0.0377171585957209
1789 0.0366032781700293
1790 0.0371354706585407
1791 0.0375363305211067
1792 0.036583931495746
1793 0.0358723911146323
1794 0.0353723391890526
1795 0.0350387779374917
1796 0.0349292767544587
1797 0.0347479619085789
1798 0.034413763632377
1799 0.0339344143867493
1800 0.0341438340644042
1801 0.0335967056453228
1802 0.0344056015213331
1803 0.0342970242102941
1804 0.0340355734030406
1805 0.0337489967544874
1806 0.0335982528825601
1807 0.0339057333767414
1808 0.0338819151123365
1809 0.0349025018513203
1810 0.0343669752279917
1811 0.035757369051377
1812 0.0349153975645701
1813 0.0342667040725549
1814 0.0351191548009714
1815 0.0344562741617362
1816 0.0364041700959206
1817 0.0366319840153058
1818 0.0381291247904301
1819 0.0383808563152949
1820 0.039172833164533
1821 0.0400391059617201
1822 0.0407133921980858
1823 0.0416531711816788
1824 0.0411180468897025
1825 0.0428926075498263
1826 0.0416191232701143
1827 0.0412460888425509
1828 0.03946861003836
1829 0.0394177995622158
1830 0.0391533486545086
1831 0.0402150476972262
1832 0.0391662903130054
1833 0.0384221573670705
1834 0.0380755973358949
1835 0.0366771072149277
1836 0.0360462752481302
1837 0.0351200414200624
1838 0.0351658177872499
1839 0.035593431442976
1840 0.0357460454106331
1841 0.0345525133113066
1842 0.0345449820160866
1843 0.0340569031735261
1844 0.0345227519671122
1845 0.0357604709764322
1846 0.0370584266881148
1847 0.0383380663891633
1848 0.0383575422068437
1849 0.0382323277493318
1850 0.037675308684508
1851 0.0359831514457862
1852 0.0357854142785072
1853 0.0349650544424852
1854 0.0355330084760984
1855 0.0358202184240023
1856 0.0350948795676231
1857 0.0344990827143192
1858 0.035018773128589
1859 0.034318661938111
1860 0.03509571403265
1861 0.0361053558687369
1862 0.0368945027391116
1863 0.0374125130474567
1864 0.0373575414220492
1865 0.037274652471145
1866 0.0372778251767159
1867 0.0367817642788092
1868 0.036587896446387
1869 0.0361283843715986
1870 0.0369358124832312
1871 0.0384087599813938
1872 0.0390824576218923
1873 0.0399971281488736
1874 0.041097030043602
1875 0.0413214899599552
1876 0.0424283035099506
1877 0.0422930307686329
1878 0.043616633862257
1879 0.0430551084379355
1880 0.0417569577693939
1881 0.0442714281380177
1882 0.0421311656634013
1883 0.0405460993448893
1884 0.040778091798226
1885 0.0381148631374041
1886 0.039490170776844
1887 0.0398999862372875
1888 0.0402161305149396
1889 0.0387468971312046
1890 0.0395637539525827
1891 0.0402010579903921
1892 0.0398118992646535
1893 0.0401631606121858
1894 0.0391472702225049
1895 0.0374248834947745
1896 0.036897249519825
1897 0.0376503554483255
1898 0.0360582694411278
1899 0.0365569802622
1900 0.0369024177392324
1901 0.0372361168265343
1902 0.0369529028733571
1903 0.0359033333758513
1904 0.0359858920176824
1905 0.0358572229743004
1906 0.0358209361632665
1907 0.0348760771254698
1908 0.03624602034688
1909 0.0359968754152457
1910 0.0354778791467349
1911 0.0363015135129293
1912 0.0351013131439686
1913 0.0367484254141649
1914 0.0360249790052573
1915 0.0367468384404977
1916 0.0373753433426221
1917 0.0369498742123445
1918 0.0357949323952198
1919 0.0382490294675032
1920 0.0365126430988312
1921 0.0364818337062995
1922 0.0369449766973654
1923 0.0348826348781586
1924 0.0350412813325723
1925 0.0364490312834581
1926 0.0346606535216173
1927 0.0350682511925697
1928 0.0353666779895624
1929 0.0336330061157544
1930 0.0354406647384167
1931 0.0359244892994563
1932 0.0347358994185925
1933 0.0349119156599045
1934 0.0355228905876478
1935 0.0336452126502991
1936 0.036103468388319
1937 0.0379126518964767
1938 0.0371061153709888
1939 0.0390059215327104
1940 0.0400740094482899
1941 0.0379178822040558
1942 0.0413161466519038
1943 0.0434064641594887
1944 0.0404349838693937
1945 0.0404018039504687
1946 0.0446461103856564
1947 0.0456789744397004
1948 0.0433709000547727
1949 0.0438445533315341
1950 0.0439405379196008
1951 0.0470656491816044
1952 0.046330997099479
1953 0.0460152601202329
1954 0.0483384008208911
1955 0.050494263569514
1956 0.0537417816619078
1957 0.0538236697514852
1958 0.0502192700902621
1959 0.0496925711631775
1960 0.0509897458056609
1961 0.0494688972830772
1962 0.0485328957438469
1963 0.0457488720615705
1964 0.0460687230030696
1965 0.0434692551692327
1966 0.0397502481937408
1967 0.0389985665678978
1968 0.0383056464294593
1969 0.0378236261506875
1970 0.0370805946489175
1971 0.0368653535842896
1972 0.0357345206042131
1973 0.0351895429193974
1974 0.0348227595289548
1975 0.0351487857600053
1976 0.0353455220659574
1977 0.0349921571711699
1978 0.0349523400266965
1979 0.0346849846343199
1980 0.0347501275440057
1981 0.0346715599298477
1982 0.0346009793380896
1983 0.0350004297991594
1984 0.0348629044989745
1985 0.0351429022848606
1986 0.0349099449813366
1987 0.0352248797814051
1988 0.035637674232324
1989 0.0356110235055288
1990 0.0357951844731967
1991 0.03509898049136
1992 0.0357600003480911
1993 0.035417137046655
1994 0.0355640115837256
1995 0.0354617151121298
1996 0.0351808145642281
1997 0.0350301750004292
1998 0.0347222040096919
1999 0.0347250886261463
2000 0.0342069764931997
};
\end{axis}

\end{tikzpicture}

%% file: onlymseplot.tex
\begin{tikzpicture}[scale = 0.55]

\definecolor{darkgray176}{RGB}{176,176,176}

\begin{axis}[
tick align=outside,
tick pos=left,
x grid style={darkgray176},
xlabel={Number of Epochs},
xmajorgrids,
xmin=-98.95, xmax=2099.95,
xtick style={color=black},
y grid style={darkgray176},
ylabel={Training Loss},
ymajorgrids,
ymin=32.7803525765737, ymax=139.471325572332,
ytick style={color=black}
]
\addplot [semithick, blue]
table {%
1 134.621735890706
2 131.652170817057
3 129.911249796549
4 128.499004364014
5 127.735392252604
6 127.024075826009
7 126.159891764323
8 125.37559000651
9 124.637280782064
10 123.892649332682
11 123.085370381673
12 122.210966746012
13 121.312187194824
14 120.375885009766
15 119.42990620931
16 118.38170115153
17 117.314151763916
18 116.225593566895
19 115.058942159017
20 113.852607727051
21 112.612747192383
22 111.260021209717
23 109.859306971232
24 108.465068817139
25 106.952529271444
26 105.39138730367
27 103.80437151591
28 102.081301371257
29 100.368423461914
30 98.5802917480469
31 96.7739950815837
32 94.8191757202148
33 92.9913756052653
34 90.9354044596354
35 88.9568583170573
36 86.8320109049479
37 84.9282976786296
38 82.8031558990479
39 80.8138656616211
40 78.7919273376465
41 76.6121012369792
42 75.002997080485
43 72.8007049560547
44 71.1476103464762
45 69.3783696492513
46 67.644282023112
47 66.1300598780314
48 64.8614972432454
49 63.5485954284668
50 62.6142374674479
51 61.7057218551636
52 60.8498999277751
53 59.6892112096151
54 58.5390949249268
55 57.5840803782145
56 56.7503579457601
57 55.4920870463053
58 55.5730247497559
59 54.6477018992106
60 54.2296940485636
61 53.498706817627
62 52.6789668401082
63 52.3620824813843
64 52.3216854731242
65 52.5530300140381
66 52.480419476827
67 51.7046378453573
68 51.7369553248088
69 51.6528317133586
70 50.1699380874634
71 49.9860178629557
72 49.6883576711019
73 49.0285739898682
74 48.4903955459595
75 48.4450079600016
76 48.0795202255249
77 47.7877775828044
78 47.2907206217448
79 46.825829188029
80 46.3768625259399
81 46.1301024754842
82 45.8641751607259
83 45.6805105209351
84 45.5015796025594
85 45.3730799357096
86 45.1289437611898
87 44.9784762064616
88 44.7408727010091
89 44.7890065511068
90 44.5494785308838
91 44.5165138244629
92 44.4173930486043
93 44.5849924087524
94 44.5854533513387
95 44.8544012705485
96 45.1886889139811
97 45.4105847676595
98 46.7723350524902
99 45.8602234522502
100 44.6293563842773
101 44.9895356496175
102 44.2958081563314
103 44.077446937561
104 43.9051275253296
105 43.6662747065226
106 43.4593353271484
107 43.1342042287191
108 42.966677347819
109 43.1600383122762
110 42.7471984227498
111 42.5289328893026
112 42.6168022155762
113 42.2279644012451
114 42.1539408365885
115 41.9303689002991
116 41.895744005839
117 41.77370262146
118 41.648942788442
119 41.5794607798258
120 41.5248619715373
121 41.3749440511068
122 41.3055884043376
123 41.2221997578939
124 41.192197004954
125 41.0953289667765
126 41.0209666887919
127 40.9799426396688
128 40.988208770752
129 40.8972020149231
130 40.8861443201701
131 41.0274381637573
132 41.0216830571493
133 40.968722820282
134 41.0389852523804
135 41.1500962575277
136 41.148553053538
137 40.9813725153605
138 40.9648429552714
139 41.2336378097534
140 41.0300296147664
141 40.8708392779032
142 41.0316019058228
143 40.6673855781555
144 40.7773323059082
145 40.7139019966125
146 40.5914999643962
147 40.6495641072591
148 40.3199903170268
149 40.6425453821818
150 40.2826730410258
151 40.4969765345256
152 40.3522020975749
153 40.1931238174438
154 40.3750693003337
155 40.0042384465536
156 40.2247344652812
157 39.8536958694458
158 39.9546192487081
159 39.7219775517782
160 39.7110873858134
161 39.63334496816
162 39.5480157534281
163 39.4994802474976
164 39.4369535446167
165 39.4559766451518
166 39.4017786979675
167 39.3676614761353
168 39.3393732706706
169 39.2898232142131
170 39.2909293174744
171 39.2624831199646
172 39.2717326482137
173 39.3306051890055
174 39.3086667060852
175 39.3981866836548
176 39.616973400116
177 39.6587182680766
178 39.7802251180013
179 40.1206720670064
180 40.363122622172
181 40.6728302637736
182 40.5440966288249
183 40.4635750452677
184 40.5705246925354
185 39.9441143671672
186 40.105073928833
187 39.6589144070943
188 39.7127606074015
189 39.525715192159
190 39.6635378201803
191 39.592485109965
192 39.3538366953532
193 39.6023383140564
194 39.2457984288534
195 39.5599160194397
196 39.1814870834351
197 39.2569233576457
198 39.1306994756063
199 39.1288735071818
200 39.0777824719747
201 38.944676399231
202 38.9121332168579
203 38.8210852940877
204 38.8193422953288
205 38.7297666867574
206 38.6894067128499
207 38.6510922114054
208 38.6289361317952
209 38.6031662623088
210 38.5774335861206
211 38.5434409777323
212 38.5530242919922
213 38.5358864466349
214 38.5236721038818
215 38.5438013076782
216 38.5257937113444
217 38.5357046127319
218 38.583119392395
219 38.5627737045288
220 38.5908915201823
221 38.6243918736776
222 38.5999410947164
223 38.6336145401001
224 38.7073926925659
225 38.705491065979
226 38.7672292391459
227 38.8023322423299
228 38.7390651702881
229 38.8761898676554
230 38.9256207148234
231 38.9731237093608
232 39.258747736613
233 39.3471021652222
234 39.5477565129598
235 40.1178515752157
236 40.1864668528239
237 40.0222188631693
238 40.5607357025146
239 39.9260729153951
240 39.6782140731812
241 39.7084426879883
242 39.3843108812968
243 39.4324873288473
244 39.0789510409037
245 39.2656784057617
246 38.92178662618
247 39.0800166130066
248 38.6901419957479
249 38.833748181661
250 38.5846254030863
251 38.6620128949483
252 38.4506068229675
253 38.5506059328715
254 38.4226136207581
255 38.4911082585653
256 38.3561367988586
257 38.422239780426
258 38.3203970591227
259 38.3919045130412
260 38.2752149899801
261 38.2937124570211
262 38.2162410418193
263 38.2329123814901
264 38.2026017506917
265 38.1796328226725
266 38.1631310780843
267 38.1704012552897
268 38.1424125035604
269 38.1286904017131
270 38.1069072087606
271 38.1066656112671
272 38.1034625371297
273 38.1163506507874
274 38.1091190973918
275 38.0974179903666
276 38.1059424082438
277 38.0928723017375
278 38.0932319959005
279 38.1076676050822
280 38.0994421641032
281 38.1384112040202
282 38.1486562093099
283 38.1944154103597
284 38.2505439122518
285 38.2279411951701
286 38.3083713849386
287 38.4293893178304
288 38.3358681996663
289 38.4146254857381
290 38.5994968414307
291 38.5271218617757
292 38.5802814165751
293 38.7065896987915
294 38.6414601008097
295 38.8388778368632
296 38.8642473220825
297 39.2763713200887
298 39.3220357894897
299 39.2715333302816
300 39.7643297513326
301 39.2262516021729
302 39.3791154225667
303 39.1039339701335
304 39.0630076726278
305 38.8154109319051
306 38.8277784983317
307 38.7757288614909
308 38.5316232045492
309 38.6303195953369
310 38.4337822596232
311 38.4026807149251
312 38.3419880867004
313 38.2409661610921
314 38.2385505040487
315 38.1495464642843
316 38.1121872266134
317 38.0772720972697
318 38.081742922465
319 38.0278517405192
320 38.0562292734782
321 38.0225068728129
322 38.0106503168742
323 38.0218470891317
324 38.012527624766
325 38.032431602478
326 38.0439510345459
327 38.0411586761475
328 38.0345226923625
329 38.0706637700399
330 38.0656099319458
331 38.0985345840454
332 38.0985420544942
333 38.0975282986959
334 38.0934933026632
335 38.1338063875834
336 38.1445436477661
337 38.1712131500244
338 38.1700541178385
339 38.140690167745
340 38.1615397135417
341 38.1709157625834
342 38.212571144104
343 38.2074184417725
344 38.2335885365804
345 38.1893157958984
346 38.2069730758667
347 38.216628074646
348 38.3019002278646
349 38.2569252649943
350 38.2834374109904
351 38.289032459259
352 38.2494738896688
353 38.3200721740723
354 38.3609093030294
355 38.3641780217489
356 38.3264266649882
357 38.4206275939941
358 38.3269027074178
359 38.4841173489889
360 38.5104223887126
361 38.6492009162903
362 38.7253142992655
363 38.7569532394409
364 38.8528685569763
365 39.1354540189107
366 39.0428980191549
367 39.1239856084188
368 39.3968985875448
369 38.9550143877665
370 39.2006060282389
371 39.1224077542623
372 38.9594815572103
373 39.0906551678975
374 38.8045293490092
375 39.1881971359253
376 38.7544501622518
377 39.0872996648153
378 38.6753403345744
379 38.9033411343892
380 38.491991519928
381 38.6875044504801
382 38.424921353658
383 38.3812996546427
384 38.225297609965
385 38.2768745422363
386 38.1552480061849
387 38.1251579920451
388 38.041384379069
389 38.0139061609904
390 37.9560813903809
391 37.950811068217
392 37.8920421600342
393 37.9005556106567
394 37.8828422228495
395 37.8696320851644
396 37.8418285051982
397 37.8475775718689
398 37.8465649286906
399 37.8436015446981
400 37.8433275222778
401 37.831480662028
402 37.8100377718608
403 37.8153338432312
404 37.7915341059367
405 37.7995351155599
406 37.7981850306193
407 37.7923205693563
408 37.7897958755493
409 37.7960316340129
410 37.7779623667399
411 37.7976806958516
412 37.806814511617
413 37.8126511573792
414 37.8185749053955
415 37.8171006838481
416 37.8172742525736
417 37.8273315429688
418 37.8319536844889
419 37.8317747116089
420 37.8489243189494
421 37.8330217997233
422 37.8707971572876
423 37.8538173039754
424 37.8627026875814
425 37.8989292780558
426 37.9109013875326
427 37.9663707415263
428 38.0427071253459
429 38.0805460611979
430 38.0634953180949
431 38.1975283622742
432 38.2580350240072
433 38.3629557291667
434 38.4319276809692
435 38.5151306788127
436 38.6458965937297
437 38.6175915400187
438 38.7204065322876
439 38.7780038515727
440 38.9586846033732
441 38.6914812723796
442 38.9087444941203
443 38.8846861521403
444 38.7137304941813
445 38.8540857632955
446 38.76824537913
447 38.4890451431274
448 38.616444905599
449 38.2569697697957
450 38.3128735224406
451 38.2740341822306
452 38.1355257034302
453 38.1986567179362
454 38.0344994862874
455 38.0505622227987
456 38.0006151199341
457 37.9564050038656
458 37.9398612976074
459 37.9329322179159
460 37.9428504308065
461 37.9334483146667
462 37.9567453066508
463 37.9409236907959
464 37.8913383483887
465 37.9598269462585
466 37.8747603098551
467 37.9398206075033
468 37.9557914733887
469 37.9303560256958
470 37.973908106486
471 37.9978869756063
472 37.9060665766398
473 38.0137518246969
474 37.9604002634684
475 38.0076187451681
476 38.0323435465495
477 38.0359999338786
478 38.0136868158976
479 38.0681522687276
480 38.0043897628784
481 38.0726973215739
482 38.0410981178284
483 38.0882352193197
484 38.0482476552327
485 38.1021194458008
486 38.0699504216512
487 38.0645033518473
488 38.0720233917236
489 38.060019493103
490 38.0637404123942
491 38.1397970517476
492 38.0423053105672
493 38.1299947102865
494 38.12908522288
495 38.019074122111
496 38.1669772466024
497 38.1393610636393
498 38.0931393305461
499 38.2611028353373
500 38.0915137926737
501 38.0591740608215
502 38.1929041544596
503 38.0394817988078
504 38.1594352722168
505 38.2000513076782
506 38.0297296841939
507 38.0842603047689
508 38.1748695373535
509 38.0933685302734
510 38.2085123062134
511 38.138444741567
512 38.0575876235962
513 38.1634581883748
514 38.262575785319
515 38.2468684514364
516 38.296275138855
517 38.2277293205261
518 38.2397521336873
519 38.3797197341919
520 38.5157267252604
521 38.5063134829203
522 38.4724235534668
523 38.485729376475
524 38.4505074818929
525 38.6372378667196
526 38.5781041781108
527 38.4490464528402
528 38.3309965133667
529 38.2319612503052
530 38.1624533335368
531 38.1519683202108
532 38.0785884857178
533 38.0911931991577
534 38.1259074211121
535 38.0659605662028
536 38.1319723129272
537 38.1228162447611
538 38.0524237950643
539 38.1305680274963
540 38.1550906499227
541 38.0948336919149
542 38.1614631017049
543 38.1512285868327
544 38.1327408154806
545 38.1150175730387
546 38.1458568572998
547 38.1569630304972
548 38.0488739013672
549 38.1029396057129
550 38.0907592773438
551 37.9958537419637
552 38.0626926422119
553 38.0352274576823
554 37.9758208592733
555 37.9988670349121
556 37.9737536112467
557 37.9067095120748
558 37.967293103536
559 37.9320619901021
560 37.8791166941325
561 37.9399658838908
562 37.9123983383179
563 37.8678801854452
564 37.993726571401
565 37.9849955240885
566 37.938507715861
567 38.0460189183553
568 38.0789643923442
569 37.9556183815002
570 38.0881301561991
571 38.1224845250448
572 38.008927822113
573 38.110297203064
574 38.1148586273193
575 38.0296335220337
576 38.1772836049398
577 38.1568692525228
578 38.2931467692057
579 38.4215408960978
580 38.4095630645752
581 38.7231969833374
582 38.8397278785706
583 38.6778004964193
584 39.1119038263957
585 38.9228677749634
586 38.8526302973429
587 39.0015996297201
588 38.7066923777262
589 38.8957204818726
590 38.5137802759806
591 38.7237246831258
592 38.551833152771
593 38.5003492037455
594 38.5878006617228
595 38.3050028483073
596 38.5664981206258
597 38.230442682902
598 38.4538876215617
599 38.1994272867839
600 38.3074259757996
601 38.1413593292236
602 38.1356294949849
603 38.0522918701172
604 37.9893023173014
605 37.9576903978984
606 37.8979549407959
607 37.8782828648885
608 37.8356828689575
609 37.7988529205322
610 37.7844909032186
611 37.7531439463298
612 37.7561483383179
613 37.7303606669108
614 37.7120354970296
615 37.7047732671102
616 37.6897819836934
617 37.6908852259318
618 37.6782393455505
619 37.6798356374105
620 37.6748797098796
621 37.6712106068929
622 37.6731532414754
623 37.6693067550659
624 37.6748178799947
625 37.6732646624247
626 37.6747109095256
627 37.6719792683919
628 37.6677765846252
629 37.6757017771403
630 37.6737062136332
631 37.6750380198161
632 37.6770741144816
633 37.6784423192342
634 37.6806348164876
635 37.6822862625122
636 37.6928935050964
637 37.7024558385213
638 37.7095546722412
639 37.7192180951436
640 37.7429324785868
641 37.745183467865
642 37.735226949056
643 37.7679233551025
644 37.7842858632406
645 37.7714802424113
646 37.8043181101481
647 37.8403876622518
648 37.8421751658122
649 37.8748254776001
650 37.9038616816203
651 37.9459673563639
652 38.0504595438639
653 38.109120686849
654 38.1746106147766
655 38.3360476493835
656 38.4920959472656
657 38.508821328481
658 38.5426457722982
659 38.8907845815023
660 38.9344034194946
661 38.7083365122477
662 38.9792617162069
663 38.8600775400798
664 38.5655074119568
665 38.6725638707479
666 38.507930914561
667 38.3889675140381
668 38.3702351252238
669 38.2345059712728
670 38.2164397239685
671 38.1660916010539
672 38.1918625831604
673 38.1843956311544
674 38.1312608718872
675 38.1987373034159
676 38.0738824208577
677 38.1702197392782
678 38.0187873840332
679 38.0785191853841
680 38.0169332822164
681 38.0010844866435
682 38.0266097386678
683 37.8915758132935
684 37.9825236002604
685 37.8638089497884
686 37.913228670756
687 37.8243842124939
688 37.8061459859212
689 37.8218302726746
690 37.7749799092611
691 37.7792812983195
692 37.7530188560486
693 37.7303740183512
694 37.7526160875956
695 37.7304743131002
696 37.728727499644
697 37.7628142038981
698 37.7327709197998
699 37.7698383331299
700 37.8010670344035
701 37.7617742220561
702 37.8495639165243
703 37.827544371287
704 37.8532683054606
705 37.9025387763977
706 37.8794186909994
707 37.9013236363729
708 37.9363660812378
709 37.8994887669881
710 37.941530863444
711 37.9443715413411
712 37.9141565958659
713 37.9545532862345
714 37.9253023465474
715 37.8942991892497
716 37.9437405268351
717 37.8864034016927
718 37.9078648885091
719 37.9331297874451
720 37.8660907745361
721 37.9213352203369
722 37.8900316556295
723 37.8696932792664
724 37.9296987851461
725 37.8793821334839
726 37.9090331395467
727 37.9519232114156
728 37.8849741617839
729 37.993067741394
730 37.964802424113
731 37.9711249669393
732 38.0829391479492
733 38.0193661053975
734 38.100746790568
735 38.2259305318197
736 38.0786867141724
737 38.3238115310669
738 38.2428922653198
739 38.2748874028524
740 38.4512612024943
741 38.2860883076986
742 38.4725777308146
743 38.4098049799601
744 38.4370632171631
745 38.5912405649821
746 38.4079597791036
747 38.6015431086222
748 38.4003481864929
749 38.5920332272848
750 38.4284639358521
751 38.515601793925
752 38.4535013834635
753 38.4143155415853
754 38.4719285964966
755 38.2330993016561
756 38.4007000923157
757 38.1727027893066
758 38.23699315389
759 38.0846761067708
760 38.0687961578369
761 38.0328489939372
762 37.9145042101542
763 37.9847812652588
764 37.8550092379252
765 37.9153744379679
766 37.8067366282145
767 37.8542873064677
768 37.7871583302816
769 37.8566559155782
770 37.783173084259
771 37.8223683039347
772 37.8252641359965
773 37.830774307251
774 37.7816632588704
775 37.7945458094279
776 37.7866856257121
777 37.8150067329407
778 37.8024396896362
779 37.758546034495
780 37.7889731725057
781 37.7548364003499
782 37.7480449676514
783 37.7567481994629
784 37.7722956339518
785 37.7473230361938
786 37.7663240432739
787 37.7621987660726
788 37.7223016421
789 37.7951013247172
790 37.7813007036845
791 37.7806468009949
792 37.8299748102824
793 37.7822259267171
794 37.8236570358276
795 37.8210229873657
796 37.8394018809001
797 37.7989962895711
798 37.8785409927368
799 37.8550345102946
800 37.8614551226298
801 37.960991859436
802 37.8188899358114
803 37.9909737904867
804 37.9171481132507
805 37.9472931226095
806 37.9685487747192
807 37.9358501434326
808 37.9763898849487
809 37.9681523640951
810 38.0877118110657
811 37.943691889445
812 38.0513954162598
813 38.0032175381978
814 37.9891204833984
815 38.1263114611308
816 37.9693196614583
817 38.1567845344543
818 38.0519936879476
819 38.0788383483887
820 38.1174303690592
821 38.1023947397868
822 38.1875948905945
823 38.0708363850911
824 38.2240519523621
825 38.0168302853902
826 38.2439839045207
827 38.1365574200948
828 38.112421353658
829 38.1826674143473
830 38.0708621342977
831 38.1167154312134
832 38.0718501408895
833 38.154699643453
834 37.9773996671041
835 38.1593300501506
836 38.0045507748922
837 38.0348459879557
838 38.0440196990967
839 37.997763633728
840 38.0338220596313
841 38.0237429936727
842 38.0381196339925
843 37.9327322642008
844 38.031409740448
845 37.9160340627035
846 37.965628306071
847 37.9737780888875
848 37.9259433746338
849 37.9348119099935
850 37.9639158248901
851 37.9122142791748
852 37.9018723169963
853 37.9147860209147
854 37.8567409515381
855 37.8829290072123
856 37.8766563733419
857 37.8748691876729
858 37.8722476959229
859 37.9564631779989
860 37.88663037618
861 37.9328055381775
862 37.9010130564372
863 37.85298426946
864 37.8580592473348
865 37.8526070912679
866 37.8384267489115
867 37.8359060287476
868 37.8865529696147
869 37.8523368835449
870 37.9311068852743
871 37.9086449940999
872 37.9109112421672
873 37.9342481295268
874 37.9473002751668
875 37.9802899360657
876 37.9798895517985
877 38.0104931195577
878 37.990203221639
879 38.1734689076742
880 38.1173858642578
881 38.185738881429
882 38.1598558425903
883 38.1839348475138
884 38.3105223973592
885 38.2047521273295
886 38.3399378458659
887 38.3162706693014
888 38.4656003316244
889 38.5619069735209
890 38.7337857882182
891 38.4863622983297
892 38.4718367258708
893 38.5137799580892
894 38.4399007161458
895 38.4249971707662
896 38.4609285990397
897 38.2789888381958
898 38.1563762029012
899 38.2494627634684
900 38.0565315882365
901 38.1023809115092
902 38.0993191401164
903 37.9815694491068
904 38.0331783294678
905 37.966100692749
906 37.882256825765
907 37.9068619410197
908 37.8765614827474
909 37.8720903396606
910 37.81503089269
911 37.8061215082804
912 37.8160875638326
913 37.7995351155599
914 37.8174041112264
915 37.8331693013509
916 37.8733531634013
917 37.8640244801839
918 37.9234116872152
919 37.8743778864543
920 37.9650921821594
921 37.9732673962911
922 38.0021810531616
923 38.0389320055644
924 37.9546372095744
925 37.9714196523031
926 37.9796317418416
927 37.9872535069784
928 37.9071466128031
929 37.9197355906169
930 37.9050849278768
931 37.8652679125468
932 37.8830509185791
933 37.8771586418152
934 37.8836139043172
935 37.8840711911519
936 37.8572869300842
937 37.8246660232544
938 37.8541393280029
939 37.830540339152
940 37.8001233736674
941 37.7895126342773
942 37.7928948402405
943 37.7941970825195
944 37.7976104418437
945 37.7854067484538
946 37.7845689455668
947 37.7772939999898
948 37.7761346499125
949 37.7828478813171
950 37.7611640294393
951 37.788506825765
952 37.7430308659871
953 37.7704043388367
954 37.7682607968648
955 37.7471714019775
956 37.738702138265
957 37.7363535563151
958 37.7602000236511
959 37.7514966328939
960 37.7720629374186
961 37.7908024787903
962 37.8111402193705
963 37.8390251795451
964 37.8463719685872
965 37.8912862141927
966 37.876189549764
967 37.8694200515747
968 37.993156115214
969 37.91401831309
970 37.9686759312948
971 38.018217086792
972 37.9935496648153
973 37.9692409833272
974 38.012705485026
975 38.1061844825745
976 38.0396690368652
977 38.1692768732707
978 38.2580302556356
979 38.2000347773234
980 38.3000588417053
981 38.3360946973165
982 38.3099706967672
983 38.2721878687541
984 38.3654400507609
985 38.1753424008687
986 38.2100156148275
987 38.1603422164917
988 38.0148259798686
989 38.1189823150635
990 37.8838237126668
991 37.9507519404093
992 37.8524068196615
993 37.8409606615702
994 37.8139619827271
995 37.7527888615926
996 37.7964893976847
997 37.7286589940389
998 37.7594467798869
999 37.716289361318
1000 37.7310984929403
1001 37.7146194775899
1002 37.7174520492554
1003 37.7326270739237
1004 37.7011426289876
1005 37.7472983996073
1006 37.7052636146545
1007 37.7802279790243
1008 37.7054484685262
1009 37.7567885716756
1010 37.7461775143941
1011 37.7311231295268
1012 37.7654705047607
1013 37.7079563140869
1014 37.762571811676
1015 37.7193654378255
1016 37.7505137125651
1017 37.7371624310811
1018 37.7532847722371
1019 37.7814219792684
1020 37.737738609314
1021 37.8132279713949
1022 37.751446723938
1023 37.7829931577047
1024 37.7954839070638
1025 37.7379032770793
1026 37.796046257019
1027 37.7341470718384
1028 37.7726667722066
1029 37.7433698972066
1030 37.7879800796509
1031 37.806500116984
1032 37.7790238062541
1033 37.9118534723918
1034 37.8281243642171
1035 37.9166332880656
1036 37.9047393798828
1037 37.8546098073324
1038 37.920293490092
1039 37.8552815119425
1040 37.9070111910502
1041 37.8979581197103
1042 37.9730965296427
1043 37.9832367897034
1044 38.064079284668
1045 38.1681381861369
1046 38.0811554590861
1047 38.2330331802368
1048 38.2078189849854
1049 38.2262268066406
1050 38.2355483373006
1051 38.1846831639608
1052 38.2843106587728
1053 38.1201985677083
1054 38.2234175999959
1055 38.2561434110006
1056 38.2078968683879
1057 38.2076431910197
1058 38.3438606262207
1059 38.1353470484416
1060 38.1492878595988
1061 38.2163804372152
1062 38.0397186279297
1063 38.0976092020671
1064 37.97993628184
1065 37.9489227930705
1066 37.8775908152262
1067 37.8706585566203
1068 37.8075731595357
1069 37.8120684623718
1070 37.8047307332357
1071 37.7496345837911
1072 37.8215386072795
1073 37.7310431798299
1074 37.7764123280843
1075 37.7616437276204
1076 37.7392856280009
1077 37.7721713383993
1078 37.7466508547465
1079 37.7215994199117
1080 37.741636912028
1081 37.6988433202108
1082 37.7038499514262
1083 37.7215433120728
1084 37.6937478383382
1085 37.7020133336385
1086 37.6930459340413
1087 37.6609002749125
1088 37.6844266255697
1089 37.6622165044149
1090 37.6752297083537
1091 37.6610476175944
1092 37.6617827415466
1093 37.6592032114665
1094 37.6527954737345
1095 37.67036596934
1096 37.6649303436279
1097 37.6656757990519
1098 37.6885353724162
1099 37.6658310890198
1100 37.6731797854106
1101 37.6839725176493
1102 37.684014638265
1103 37.6970715522766
1104 37.7090520858765
1105 37.6999835968018
1106 37.6981355349223
1107 37.722691377004
1108 37.7294238408407
1109 37.7493543624878
1110 37.7696266174316
1111 37.781909942627
1112 37.8017724355062
1113 37.8548997243245
1114 37.914280573527
1115 37.8745673497518
1116 37.9175680478414
1117 37.9657796223958
1118 37.9833701451619
1119 37.9769868850708
1120 38.0304228464762
1121 38.0166524251302
1122 37.9877446492513
1123 38.0626811981201
1124 38.0716285705566
1125 38.036597887675
1126 38.1512171427409
1127 38.2296592394511
1128 38.1979697545369
1129 38.2565495173136
1130 38.366117477417
1131 38.3503886858622
1132 38.2937599817912
1133 38.4176565806071
1134 38.3982712427775
1135 38.2517267862956
1136 38.3373804092407
1137 38.2963857650757
1138 38.2025632858276
1139 38.3701982498169
1140 38.2679316202799
1141 38.2523155212402
1142 38.3319883346558
1143 38.2858012517293
1144 38.2795937856038
1145 38.3357942899068
1146 38.2312366167704
1147 38.2768872578939
1148 38.2784342765808
1149 38.2314988772074
1150 38.231097539266
1151 38.2248134613037
1152 38.20631980896
1153 38.2072518666585
1154 38.2121696472168
1155 38.2290538152059
1156 38.2058140436808
1157 38.2479244867961
1158 38.2140611012777
1159 38.2525863647461
1160 38.1839330991109
1161 38.1949234008789
1162 38.1845792134603
1163 38.1537637710571
1164 38.2044823964437
1165 38.0928284327189
1166 38.1858108838399
1167 38.0710023244222
1168 38.186491171519
1169 38.0826692581177
1170 38.1488231023153
1171 38.0789643923442
1172 38.083979288737
1173 38.1102619171143
1174 38.0105555852254
1175 38.0342575709025
1176 37.9133030573527
1177 37.9754034678141
1178 37.8924805323283
1179 37.8942532539368
1180 37.8145049413045
1181 37.844121615092
1182 37.806342124939
1183 37.8062423070272
1184 37.7544190088908
1185 37.7772089640299
1186 37.7670070330302
1187 37.7861413955688
1188 37.7587906519572
1189 37.7477842966715
1190 37.7329998016357
1191 37.7389497756958
1192 37.6987846692403
1193 37.6926593780518
1194 37.6997184753418
1195 37.6861893335978
1196 37.6895613670349
1197 37.6716860135396
1198 37.66055504481
1199 37.6665274302165
1200 37.6659739812215
1201 37.6586484909058
1202 37.6710410118103
1203 37.6610946655273
1204 37.6504468917847
1205 37.6565969785055
1206 37.6427628199259
1207 37.6561430295308
1208 37.6633507410685
1209 37.657320022583
1210 37.664531548818
1211 37.6629169782003
1212 37.6395789782206
1213 37.64781665802
1214 37.6463147799174
1215 37.6498003005981
1216 37.6342851320902
1217 37.6400569279989
1218 37.6331221262614
1219 37.6456165313721
1220 37.6447525024414
1221 37.6395540237427
1222 37.6455629666646
1223 37.6299422581991
1224 37.6504786809286
1225 37.639808177948
1226 37.6384496688843
1227 37.6466584205627
1228 37.6365769704183
1229 37.6440060933431
1230 37.6412698427836
1231 37.641658782959
1232 37.6555833816528
1233 37.6423873901367
1234 37.6672347386678
1235 37.6561427116394
1236 37.6645221710205
1237 37.68581310908
1238 37.667195002238
1239 37.6963515281677
1240 37.6879463195801
1241 37.6893127759298
1242 37.7139868736267
1243 37.7034791310628
1244 37.7373671531677
1245 37.7168957392375
1246 37.7442520459493
1247 37.7608064015706
1248 37.7509380976359
1249 37.8163197835286
1250 37.777556737264
1251 37.8073199590047
1252 37.8263562520345
1253 37.8255597750346
1254 37.9026927947998
1255 37.8588393529256
1256 37.9339240392049
1257 37.9097065925598
1258 37.9514923095703
1259 38.0175937016805
1260 38.0141318639119
1261 38.1023448308309
1262 38.0628457069397
1263 38.2272946039836
1264 38.1976006825765
1265 38.1438716252645
1266 38.2387440999349
1267 38.1027417182922
1268 38.1517413457235
1269 38.1153060595194
1270 38.0980799992879
1271 38.0057655970256
1272 38.0350766181946
1273 38.0019210179647
1274 37.9818483988444
1275 37.9616780281067
1276 37.9766991933187
1277 37.984824180603
1278 38.0121742884318
1279 37.9440983136495
1280 38.0185486475627
1281 38.0324897766113
1282 38.0568110148112
1283 38.1464509963989
1284 38.1805226008097
1285 38.3418242136637
1286 38.3806575139364
1287 38.438002427419
1288 38.4400240580241
1289 38.704662322998
1290 38.5741726557414
1291 38.678505897522
1292 38.6538106600443
1293 38.4633099238078
1294 38.60813331604
1295 38.3035734494527
1296 38.3783281644185
1297 38.0930859247843
1298 38.276494661967
1299 38.0173269907633
1300 38.1096410751343
1301 37.9986902872721
1302 37.9623095194499
1303 37.9881342252096
1304 37.9212576548258
1305 37.9388421376546
1306 37.8612219492594
1307 37.8515316645304
1308 37.8077907562256
1309 37.7872444788615
1310 37.762477238973
1311 37.7514497439067
1312 37.7422100702922
1313 37.724992275238
1314 37.7308815320333
1315 37.7299443880717
1316 37.729375521342
1317 37.7214272816976
1318 37.6983890533447
1319 37.7304821014404
1320 37.7296883265177
1321 37.732658068339
1322 37.6888446807861
1323 37.7018547058105
1324 37.7142238616943
1325 37.7270434697469
1326 37.6745270093282
1327 37.6862902641296
1328 37.6798453330994
1329 37.6764605840047
1330 37.6537545522054
1331 37.6452379226685
1332 37.65624888738
1333 37.6490468978882
1334 37.6566338539124
1335 37.6575406392415
1336 37.6635654767354
1337 37.666792233785
1338 37.6954468091329
1339 37.7089850107829
1340 37.7123753229777
1341 37.7224739392598
1342 37.6985152562459
1343 37.694904645284
1344 37.6755584081014
1345 37.6810609499613
1346 37.6559171676636
1347 37.6624113718669
1348 37.6500539779663
1349 37.6651436487834
1350 37.7110493977865
1351 37.7161385218302
1352 37.7540267308553
1353 37.7725989023844
1354 37.8153462409973
1355 37.7938715616862
1356 37.7752262751261
1357 37.8150666554769
1358 37.7981837590536
1359 37.7892026901245
1360 37.7919457753499
1361 37.7546319961548
1362 37.778501033783
1363 37.8353786468506
1364 37.8527266184489
1365 37.9453337987264
1366 37.987598101298
1367 38.0137934684753
1368 38.083314259847
1369 38.122358640035
1370 38.1876730918884
1371 38.2342670758565
1372 38.2703170776367
1373 38.3699185053507
1374 38.2584139506022
1375 38.4727584520976
1376 38.275982538859
1377 38.2411842346191
1378 38.3450573285421
1379 38.1223012606303
1380 38.1734441121419
1381 37.9795850118001
1382 38.0054610570272
1383 37.9537846247355
1384 37.8897652626038
1385 37.8861312866211
1386 37.9091510772705
1387 37.9505160649618
1388 37.9004238446554
1389 37.9055625597636
1390 37.9349411328634
1391 37.9259287516276
1392 37.9345153172811
1393 37.912096341451
1394 37.9017539024353
1395 37.8755346934001
1396 37.8707494735718
1397 37.906218846639
1398 37.851421991984
1399 37.9201707839966
1400 37.9039637247721
1401 37.9508681297302
1402 38.0514183044434
1403 37.9659682909648
1404 38.1104132334391
1405 38.0566498438517
1406 38.2260578473409
1407 38.1758975982666
1408 38.276287873586
1409 38.3151756922404
1410 38.2765464782715
1411 38.4476903279622
1412 38.3016017278036
1413 38.5224231084188
1414 38.3135658899943
1415 38.5347115198771
1416 38.2309401830037
1417 38.4827295939128
1418 38.2786283493042
1419 38.4325923919678
1420 38.1163180669149
1421 38.2348942756653
1422 38.0731182098389
1423 38.0956786473592
1424 37.9977836608887
1425 37.9746349652608
1426 37.9123058319092
1427 37.9346788724264
1428 37.907709757487
1429 37.896052201589
1430 37.8310527801514
1431 37.806142171224
1432 37.8043435414632
1433 37.8137219746908
1434 37.7802538871765
1435 37.7737552324931
1436 37.7268328666687
1437 37.739444732666
1438 37.7319142023722
1439 37.7237739562988
1440 37.7257641156515
1441 37.6931338310242
1442 37.7122707366943
1443 37.705249786377
1444 37.7090169588725
1445 37.6977860132853
1446 37.696916103363
1447 37.6930433909098
1448 37.6967763900757
1449 37.6917848587036
1450 37.7140231132507
1451 37.7005319595337
1452 37.705296198527
1453 37.6968390146891
1454 37.7135082880656
1455 37.7135442097982
1456 37.7202860514323
1457 37.7322942415873
1458 37.7366609573364
1459 37.7186438242594
1460 37.7349785168966
1461 37.7534265518188
1462 37.7845770517985
1463 37.7547690073649
1464 37.7770983378092
1465 37.8111352920532
1466 37.842843691508
1467 37.8458684285482
1468 37.8503459294637
1469 37.8906173706055
1470 37.9318502744039
1471 37.89057970047
1472 37.9500096638997
1473 37.9891080856323
1474 37.9396540323893
1475 37.9989775021871
1476 38.0282729466756
1477 37.9835074742635
1478 38.0579775174459
1479 38.0743198394775
1480 38.0119582811991
1481 38.0662027994792
1482 38.0849752426147
1483 38.0060377120972
1484 38.0295990308126
1485 38.0404332478841
1486 37.9590622584025
1487 37.9459292093913
1488 37.9538129170736
1489 37.8433647155762
1490 37.8856062889099
1491 37.8434378306071
1492 37.7645729382833
1493 37.8026013374329
1494 37.7421766916911
1495 37.7427347501119
1496 37.7604014078776
1497 37.6911239624023
1498 37.6962575912476
1499 37.6865674654643
1500 37.6750834782918
1501 37.6872545878092
1502 37.6763860384623
1503 37.6716731389364
1504 37.6836916605631
1505 37.6714495023092
1506 37.66734790802
1507 37.6840271949768
1508 37.6643346150716
1509 37.6773902575175
1510 37.694215297699
1511 37.6915095647176
1512 37.7066555023193
1513 37.7133143742879
1514 37.7046798070272
1515 37.7083915074666
1516 37.732040087382
1517 37.735807577769
1518 37.7332010269165
1519 37.7576815287272
1520 37.72696129481
1521 37.754633585612
1522 37.7675526936849
1523 37.7412773768107
1524 37.7721665700277
1525 37.7634563446045
1526 37.7360547383626
1527 37.7695153554281
1528 37.7648765246073
1529 37.7502857844035
1530 37.7919394175212
1531 37.7644999821981
1532 37.7819849650065
1533 37.7927716573079
1534 37.772744178772
1535 37.8053523699443
1536 37.7982433636983
1537 37.7959489822388
1538 37.8408104578654
1539 37.8110755284627
1540 37.8195948600769
1541 37.8350197474162
1542 37.8275682131449
1543 37.7981093724569
1544 37.8414754867554
1545 37.8032560348511
1546 37.7959914207458
1547 37.8373829523722
1548 37.7564274470011
1549 37.8362135887146
1550 37.8283457756042
1551 37.7489468256632
1552 37.8706620534261
1553 37.7835108439128
1554 37.8398005167643
1555 37.8870226542155
1556 37.8069785435994
1557 37.9101963043213
1558 37.8580643335978
1559 37.880473613739
1560 37.8759117126465
1561 37.824068069458
1562 37.8502348264058
1563 37.7775925000509
1564 37.8069535891215
1565 37.7723280588786
1566 37.7614854176839
1567 37.7759545644124
1568 37.7254797617594
1569 37.7381715774536
1570 37.7151521046956
1571 37.715158144633
1572 37.7200813293457
1573 37.7198289235433
1574 37.7402629852295
1575 37.743586063385
1576 37.7377958297729
1577 37.7600034077962
1578 37.7171521186829
1579 37.7533203760783
1580 37.7516562143962
1581 37.7342297236125
1582 37.7925138473511
1583 37.8107687632243
1584 37.7988176345825
1585 37.8025895754496
1586 37.8054620424906
1587 37.7824678421021
1588 37.8280022939046
1589 37.8667187690735
1590 37.8948464393616
1591 38.0135539372762
1592 38.0391546885173
1593 38.0557578404744
1594 38.1170166333516
1595 38.146710395813
1596 38.1176350911458
1597 38.189649105072
1598 38.1871379216512
1599 38.1735610961914
1600 38.2115513483683
1601 38.1757853825887
1602 38.1674962043762
1603 38.2261498769124
1604 38.091373761495
1605 38.1700188318888
1606 38.016064008077
1607 38.0873715082804
1608 37.9929917653402
1609 37.9759429295858
1610 37.998675664266
1611 37.946143468221
1612 38.0260699590047
1613 37.9774092038473
1614 38.1586256027222
1615 38.0454991658529
1616 38.3009599049886
1617 38.1646334330241
1618 38.3563914299011
1619 38.3372948964437
1620 38.314716498057
1621 38.4547174771627
1622 38.3157970110575
1623 38.6425579388936
1624 38.26340564092
1625 38.7313033739726
1626 38.2704610824585
1627 38.7560103734334
1628 38.2302783330282
1629 38.6314274470011
1630 38.2331597010295
1631 38.5056872367859
1632 38.2477300961812
1633 38.3461529413859
1634 38.1919984817505
1635 38.1689437230428
1636 38.1811668078105
1637 38.0493528048197
1638 38.0459661483765
1639 37.9348255793254
1640 37.9298725128174
1641 37.9150403340658
1642 37.9016882578532
1643 37.9216769536336
1644 37.8842080434163
1645 37.815004825592
1646 37.8471981684367
1647 37.8438215255737
1648 37.8715152740479
1649 37.8011779785156
1650 37.8337059020996
1651 37.8218590418498
1652 37.8117275238037
1653 37.9537382125854
1654 37.9725751876831
1655 38.0023930867513
1656 37.8626515070597
1657 37.9016955693563
1658 38.0177402496338
1659 38.0443585713704
1660 38.0036703745524
1661 37.9352540969849
1662 37.9868392944336
1663 38.043820699056
1664 38.0640640258789
1665 38.1007798512777
1666 38.0959335962931
1667 38.1756601333618
1668 38.2399921417236
1669 38.3211560249329
1670 38.4292910893758
1671 38.5058981577555
1672 38.749228477478
1673 38.5981291135152
1674 38.515367825826
1675 38.48495276769
1676 38.4302792549133
1677 38.3086522420247
1678 38.1701828638713
1679 38.0909288724264
1680 38.0029729207357
1681 38.0079719225566
1682 37.9659220377604
1683 37.9408610661825
1684 37.9049615859985
1685 37.8410943349202
1686 37.8033526738485
1687 37.7622566223145
1688 37.7335821787516
1689 37.734699567159
1690 37.7022345860799
1691 37.7153313954671
1692 37.6878552436829
1693 37.6860225995382
1694 37.6964154243469
1695 37.6735164324443
1696 37.6755263010661
1697 37.6658563613892
1698 37.6757678985596
1699 37.6470623016357
1700 37.6469570795695
1701 37.6577577590942
1702 37.6418944994609
1703 37.6516617139181
1704 37.6562252044678
1705 37.6494201024373
1706 37.6442543665568
1707 37.650471051534
1708 37.6524508794149
1709 37.6393211682638
1710 37.6474806467692
1711 37.64857006073
1712 37.6587562561035
1713 37.6438461939494
1714 37.6517895062765
1715 37.6473827362061
1716 37.6383600234985
1717 37.653151512146
1718 37.6502828598022
1719 37.6544683774312
1720 37.6367427508036
1721 37.6481796900431
1722 37.6620575586955
1723 37.6494935353597
1724 37.6765955289205
1725 37.668227036794
1726 37.6783839861552
1727 37.6659175554911
1728 37.6692444483439
1729 37.6959454218547
1730 37.6857074101766
1731 37.6831541061401
1732 37.7099339167277
1733 37.716116587321
1734 37.6973711649577
1735 37.7210607528687
1736 37.7592585881551
1737 37.7520484924316
1738 37.7696743011475
1739 37.7856170336405
1740 37.804879506429
1741 37.7870035171509
1742 37.7907206217448
1743 37.8279746373494
1744 37.7885513305664
1745 37.7881100972494
1746 37.828600247701
1747 37.795611222585
1748 37.7811009089152
1749 37.8315291404724
1750 37.8016103108724
1751 37.7778326670329
1752 37.8435948689779
1753 37.8075806299845
1754 37.7960443496704
1755 37.8629662195841
1756 37.8461627960205
1757 37.8217086791992
1758 37.8652130762736
1759 37.8511047363281
1760 37.8085260391235
1761 37.8405485153198
1762 37.8584262530009
1763 37.8027346928914
1764 37.8466828664144
1765 37.8442753156026
1766 37.7843370437622
1767 37.8116013209025
1768 37.8352931340536
1769 37.8024619420369
1770 37.857163588206
1771 37.8727496465047
1772 37.8370391527812
1773 37.8767218589783
1774 37.9017868041992
1775 37.8813333511353
1776 37.9329868952433
1777 37.9422591527303
1778 37.9235631624858
1779 37.956303914388
1780 37.9505974451701
1781 37.9322059949239
1782 37.9628214836121
1783 37.9278508822123
1784 37.9083630243937
1785 37.9232927958171
1786 37.8967291514079
1787 37.8847557703654
1788 37.8969167073568
1789 37.8445065816244
1790 37.8572236696879
1791 37.8344645500183
1792 37.8135646184286
1793 37.8389220237732
1794 37.7899209658305
1795 37.7944717407227
1796 37.7619110743205
1797 37.779320081075
1798 37.7634995778402
1799 37.7619191805522
1800 37.7632884979248
1801 37.7437566121419
1802 37.7894838651021
1803 37.7455209096273
1804 37.8005619049072
1805 37.7762263615926
1806 37.8143318494161
1807 37.8372624715169
1808 37.8458321889242
1809 37.9138525327047
1810 37.8718576431274
1811 37.9659517606099
1812 37.9155708948771
1813 38.0091854731242
1814 37.9841648737589
1815 37.9994446436564
1816 38.0297037760417
1817 37.955189704895
1818 38.0637458165487
1819 37.9360869725545
1820 38.0721864700317
1821 37.9165379206339
1822 38.0198435783386
1823 37.9214839935303
1824 37.9674011866252
1825 37.9225959777832
1826 37.8775539398193
1827 37.9325423240662
1828 37.8439933458964
1829 37.9479999542236
1830 37.8003142674764
1831 37.9424629211426
1832 37.8167012532552
1833 37.9389473597209
1834 37.8076488176982
1835 37.9104285240173
1836 37.8494766553243
1837 37.9127016067505
1838 37.8480806350708
1839 37.860232035319
1840 37.8639329274495
1841 37.8486220041911
1842 37.8443921407064
1843 37.781267007192
1844 37.8544626235962
1845 37.7626523971558
1846 37.828506787618
1847 37.7255132993062
1848 37.8292640050252
1849 37.7540788650513
1850 37.8468999862671
1851 37.7519925435384
1852 37.8619009653727
1853 37.8233156204224
1854 37.8771568934123
1855 37.806653658549
1856 37.8724187215169
1857 37.8381439844767
1858 37.8851027488708
1859 37.8088167508443
1860 37.8427260716756
1861 37.7971192995707
1862 37.8234634399414
1863 37.7723239262899
1864 37.7870276769002
1865 37.7582766215006
1866 37.7963097890218
1867 37.7928708394369
1868 37.8041114807129
1869 37.8013888994853
1870 37.8527046839396
1871 37.8725833892822
1872 37.8723443349202
1873 37.8460887273153
1874 37.8902571996053
1875 37.8658307393392
1876 37.8437728881836
1877 37.7896890640259
1878 37.8070100148519
1879 37.7806289990743
1880 37.7644259134928
1881 37.7522118886312
1882 37.7497029304504
1883 37.7505447069804
1884 37.7752736409505
1885 37.7556873957316
1886 37.8020180066427
1887 37.7682902018229
1888 37.8244379361471
1889 37.7574952443441
1890 37.7860717773438
1891 37.7521017392476
1892 37.8137103716532
1893 37.7598195075989
1894 37.7861032485962
1895 37.7696916262309
1896 37.799694220225
1897 37.7669674555461
1898 37.8027642567952
1899 37.7789947191874
1900 37.8318158785502
1901 37.7560386657715
1902 37.8114172617594
1903 37.7741746902466
1904 37.8060935338338
1905 37.784218788147
1906 37.7987661361694
1907 37.7674833933512
1908 37.8199248313904
1909 37.8258587519328
1910 37.8227011362712
1911 37.9038368860881
1912 37.9123570124308
1913 37.9343487421672
1914 38.0049626032511
1915 38.0525773366292
1916 37.9993336995443
1917 38.0809907913208
1918 38.0308594703674
1919 38.0057853062948
1920 38.0364678700765
1921 37.9556733767192
1922 38.0089982350667
1923 38.0092538197835
1924 37.943515141805
1925 38.0371704101562
1926 37.959939956665
1927 38.0310649871826
1928 38.0385341644287
1929 37.9807516733805
1930 38.0962468783061
1931 38.0164346694946
1932 38.1136843363444
1933 38.0848894119263
1934 38.1267096201579
1935 38.101495107015
1936 38.1135629018148
1937 38.0990648269653
1938 38.0866632461548
1939 38.0293211936951
1940 38.0981810887655
1941 38.0147151947021
1942 38.0797567367554
1943 37.9777698516846
1944 38.0265054702759
1945 37.9181067148844
1946 37.9313141504923
1947 37.8956623077393
1948 37.9330832163493
1949 37.8472011884054
1950 37.9270594914754
1951 37.8316291173299
1952 37.8593792915344
1953 37.8147246042887
1954 37.8008019129435
1955 37.7691303888957
1956 37.7673885027568
1957 37.8022572199504
1958 37.7484987576803
1959 37.8119157155355
1960 37.7954225540161
1961 37.8165868123372
1962 37.7914721171061
1963 37.7846676508586
1964 37.7740589777629
1965 37.7393957773844
1966 37.7909843126933
1967 37.7501009305318
1968 37.8060035705566
1969 37.7790142695109
1970 37.8207270304362
1971 37.7980588277181
1972 37.849289894104
1973 37.8267946243286
1974 37.8459456761678
1975 37.8585618336995
1976 37.8414959907532
1977 37.8676439921061
1978 37.8378241856893
1979 37.8673475583394
1980 37.8391316731771
1981 37.8408654530843
1982 37.8762528101603
1983 37.8496087392171
1984 37.8566185633341
1985 37.8173631032308
1986 37.8096078236898
1987 37.8005887667338
1988 37.7883121172587
1989 37.759388923645
1990 37.7420476277669
1991 37.7320510546366
1992 37.7156044642131
1993 37.7186169624329
1994 37.6815757751465
1995 37.7111639976501
1996 37.6711082458496
1997 37.6942440668742
1998 37.6846532821655
1999 37.7003507614136
2000 37.6968421936035
};
\end{axis}

\end{tikzpicture}

%% file: iclr2025_conference.bbl
\begin{thebibliography}{4}
\providecommand{\natexlab}[1]{#1}
\providecommand{\url}[1]{\texttt{#1}}
\expandafter\ifx\csname urlstyle\endcsname\relax
  \providecommand{\doi}[1]{doi: #1}\else
  \providecommand{\doi}{doi: \begingroup \urlstyle{rm}\Url}\fi

\bibitem[Agrawal et~al.(2018)Agrawal, Verschueren, Diamond, and Boyd]{agrawal2018rewriting}
Akshay Agrawal, Robin Verschueren, Steven Diamond, and Stephen Boyd.
\newblock A rewriting system for convex optimization problems.
\newblock \emph{Journal of Control and Decision}, 5\penalty0 (1):\penalty0 42--60, 2018.

\bibitem[Boyd \& Vandenberghe(2004)Boyd and Vandenberghe]{Boyd_Vandenberghe_2004}
Stephen Boyd and Lieven Vandenberghe.
\newblock \emph{Convex Optimization}.
\newblock Cambridge University Press, 2004.

\bibitem[Diamond \& Boyd(2016)Diamond and Boyd]{diamond2016cvxpy}
Steven Diamond and Stephen Boyd.
\newblock {CVXPY}: {A} {P}ython-embedded modeling language for convex optimization.
\newblock \emph{Journal of Machine Learning Research}, 17\penalty0 (83):\penalty0 1--5, 2016.

\bibitem[Femine(2024)]{femine2024kktinformedneuralnetwork}
Carmine~Delle Femine.
\newblock {KKT-Informed Neural Network}, 2024.
\newblock URL \url{https://arxiv.org/abs/2409.09087}.

\end{thebibliography}
